\documentclass{article}

\usepackage[final,nonatbib]{neurips_2021}

\usepackage[utf8]{inputenc} % allow utf-8 input
\usepackage[T1]{fontenc}    % use 8-bit T1 fonts
\usepackage{hyperref}       % hyperlinks
\usepackage{url}            % simple URL typesetting
\usepackage{booktabs}       % professional-quality tables
\usepackage{amsfonts}       % blackboard math symbols
\usepackage{nicefrac}       % compact symbols for 1/2, etc.
\usepackage{microtype}      % microtypography
\usepackage{graphicx}
\usepackage{amsmath,amssymb} % define this before the line numbering.
\usepackage{color}
\usepackage{epsfig}
\usepackage{tabularx}
\usepackage{booktabs}
\usepackage{multirow}
\usepackage{array}
\usepackage{xspace}
\usepackage{balance}
\usepackage{enumitem}
\usepackage{mathtools}
\usepackage{rotating}
\usepackage{tabulary}
\usepackage[table,dvipsnames]{xcolor}
\usepackage{subcaption}
\usepackage[export]{adjustbox}
\usepackage{ctable}
\usepackage{diagbox}
\usepackage{subcaption}
\usepackage{makecell}
\usepackage[linesnumbered,ruled]{algorithm2e}
\usepackage{bbm}
\usepackage{wrapfig}

\def\Vec#1{{\boldsymbol{#1}}}

\newcommand{\ie}{\textit{i}.\textit{e}.}
\newcommand{\eg}{\textit{e}.\textit{g}.}
\newcommand{\etal}{\textit{et al}.}

\graphicspath{{./figs/}}

\title{Align before Fuse: Vision and Language Representation Learning with Momentum Distillation
} % Replace with your title

\author{Junnan Li, ~Ramprasaath R. Selvaraju, ~Akhilesh D. Gotmare \\ \textbf{Shafiq Joty, ~Caiming Xiong, ~Steven C.H. Hoi}\\
	Salesforce Research\\
	\texttt{\{junnan.li,rselvaraju,akhilesh.gotmare,sjoty,shoi\}@salesforce.com}
}   

\begin{document}

	\maketitle
\vspace{-4ex}
   \begin{abstract}
\vspace{-1.5ex}
Large-scale vision and language representation learning has shown promising improvements on various vision-language tasks.
Most existing methods employ a transformer-based multimodal encoder to jointly model visual tokens (region-based image features) and word tokens.
Because the visual tokens and word tokens are unaligned,
it is challenging for the multimodal encoder to learn image-text interactions. 
% \rp{it would be good to mention briefly here why using a single multimodal encoder is challenging}
In this paper,
we introduce a contrastive loss to ALign the image and text representations 
BEfore Fusing (ALBEF) them through cross-modal attention,
which enables more grounded vision and language representation learning.
Unlike most existing methods, our method does not require bounding box annotations nor high-resolution images.
To improve learning from noisy web data,
we propose momentum distillation,
a self-training method which learns from pseudo-targets produced by a momentum model.
We provide a theoretical analysis of ALBEF from a mutual information maximization perspective,
showing that different training tasks can be interpreted as different ways to generate views for an image-text pair.
ALBEF achieves state-of-the-art performance on multiple downstream vision-language tasks.
On image-text retrieval, ALBEF outperforms methods that are pre-trained on orders of magnitude larger datasets.
On VQA and NLVR$^2$, ALBEF achieves absolute improvements of $2.37\%$ and $3.84\%$ compared to the state-of-the-art, while enjoying faster inference speed.  
Code and models are available at 
\textcolor{magenta}{\url{https://github.com/salesforce/ALBEF}}.
\end{abstract}
	
   \vspace{-3ex}
\section{Introduction}
\label{sec:introduction}
\vspace{-1.5ex}
Vision-and-Language Pre-training (VLP) aims to learn multimodal representations from large-scale image-text pairs that can improve downstream Vision-and-Language (V+L) tasks. 
Most existing VLP methods (\eg~LXMERT~\cite{LXMERT}, UNITER~\cite{uniter}, OSCAR~\cite{oscar}) rely on pre-trained object detectors to extract region-based image features, 
and employ a multimodal encoder to fuse the image features with word tokens.
The multimodal encoder is trained to solve tasks that require joint understanding of image and text, such as masked language modeling (MLM) and image-text matching (ITM).

While effective, this VLP framework suffers from several key limitations: (1) The image features and the word token embeddings reside in their own spaces, which makes it challenging for the multimodal encoder to learn to model their interactions; (2) The object detector is both annotation-expensive and compute-expensive, because it requires bounding box annotations during pre-training, and high-resolution (\eg~600$\times$1000) images during inference; (3) The widely used image-text datasets~\cite{CC,sbu} are collected from the web and are inherently noisy,
and existing pre-training objectives such as MLM may overfit to the noisy text and degrade the model's generalization performance.

We propose ALign BEfore Fuse (ALBEF), a new VLP framework to address these limitations.
We first encode the image and text independently with a detector-free image encoder and a text encoder.
Then we use a multimodal encoder to fuse the image features with the text features through cross-modal attention.
We introduce an intermediate image-text contrastive (ITC) loss on representations from the unimodal encoders,
which serves three purposes:
(1) it aligns the image features and the text features, making it easier for the multimodal encoder to perform cross-modal learning;
(2) it improves the unimodal encoders to better understand the semantic meaning of images and texts;
(3) it learns a common low-dimensional space to embed images and texts, which enables the image-text matching objective to find more informative samples through our contrastive hard negative mining.

To improve learning under noisy supervision,
we propose Momentum Distillation (MoD), a simple method which enables the model to leverage a larger uncurated web dataset.
During training,
we keep a momentum version of the model by taking the moving-average of its parameters, %~\cite{moco, mean_teacher}.
and use the momentum model to generate pseudo-targets as additional supervision.
With MoD, the model is not penalized for producing other reasonable outputs that are different from the web annotation.
We show that MoD not only improves pre-training,
but also downstream tasks with clean annotations.

% Contrastive learning has been shown to maximize a lower bound on mutual information between different views of an input~\cite{CPC}.
We provide theoretical justifications on ALBEF from the perspective of mutual information maximization.
Specifically,
we show that ITC and MLM maximize a lower bound on the mutual information between different views of an image-text pair,
where the views are generated by taking partial information from each pair.
From this perspective,
our momentum distillation can be interpreted as generating new views with semantically similar samples.
% different from the original image-text pair.
Therefore,
ALBEF learns vision-language representations that are invariant to semantic-preserving transformations.

We demonstrate the effectiveness of ALBEF on various downstream V+L tasks including image-text retrieval, visual question answering, visual reasoning, visual entailment, and weakly-supervised visual grounding.
ALBEF achieves substantial improvements over existing state-of-the-art methods.
On image-text retrieval, it outperforms methods that are pre-trained on orders of magnitude larger datasets (CLIP~\cite{clip} and ALIGN~\cite{align}).
On VQA and NLVR$^2$, it achieves absolute improvements of $2.37\%$ and $3.84\%$ compared to the state-of-the-art method VILLA~\cite{villa}, while enjoying much faster inference speed.  
We also provide quantitative and qualitative analysis on ALBEF using Grad-CAM~\cite{gradcam}, which reveals its ability to perform accurate object, attribute and relationship grounding implicitly.

   \vspace{-1.5ex}
\section{Related Work}
\label{sec:literature}
\vspace{-1.2ex}
\subsection{Vision-Language Representation Learning}
\vspace{-1ex}
Most existing work on vision-language representation learning fall into two categories.
The first category focuses on modelling the interactions between image and text features with transformer-based multimodal encoders~\cite{VL-BERT,ViLBERT,12in1,VisualBERT,LXMERT,ImageBERT,unicoder,uniter,oscar,ernie,villa,vinvl,soho}.
Methods in this category achieve superior performance on downstream V+L tasks that require complex reasoning over image and text (\eg~NLVR$^2$~\cite{NLVR}, VQA~\cite{VQA}),
but most of them require high-resolution input images and pre-trained object detectors.
A recent method~\cite{ViLT} improves inference speed by removing the object detector, but results in lower performance.
The second category focuses on learning separate unimodal encoders for image and text~\cite{VSE, VSR, clip, align}.
The recent CLIP~\cite{clip} and ALIGN~\cite{align} perform pre-training on massive noisy web data using a contrastive loss,
one of the most effective loss for representation learning~\cite{moco,simclr,PCL,mopro}.  
They achieve remarkable performance on image-text retrieval tasks, but lack the ability to model more complex interactions between image and text for other V+L tasks~\cite{ViLT}.

ALBEF unifies the two categories, leading to strong unimodal and multimodal representations with superior performance on both retrieval and reasoning tasks.
Furthermore, ALBEF does not require object detectors, a major computation bottleneck for many existing methods~\cite{LXMERT,uniter,oscar,villa,vinvl}.

\vspace{-1.2ex}
\subsection{Knowledge Distillation}
\vspace{-1ex}
Knowledge distillation~\cite{kd_hinton} aims to improve a student model's performance by distilling knowledge from a teacher model,
usually through matching the student's prediction with the teacher's.
While most methods focus on distilling knowledge from a pre-trained teacher model~\cite{kd_hinton,kd_att,born_again,deit, distilbert},
online distillation~\cite{DML, co_distill} simultaneously trains multiple models and use their ensemble as the teacher.
Our momentum distillation can be interpreted as a form of online self-distillation, where a temporal ensemble of the student model is used as the teacher.
Similar ideas have been explored in semi-supervised learning~\cite{mean_teacher}, label noise learning~\cite{dividemix},
and very recently in contrastive learning~\cite{self_distill}.
Different from existing studies, we theoretically and experimentally show that momentum distillation is a generic learning algorithm that can improve the model's performance on many V+L tasks.

   \vspace{-1.5ex}
\section{ALBEF Pre-training}
\label{sec:method}
\vspace{-1.2ex}
\begin{figure*}[!t]

	\centering
	\includegraphics[width=\textwidth]{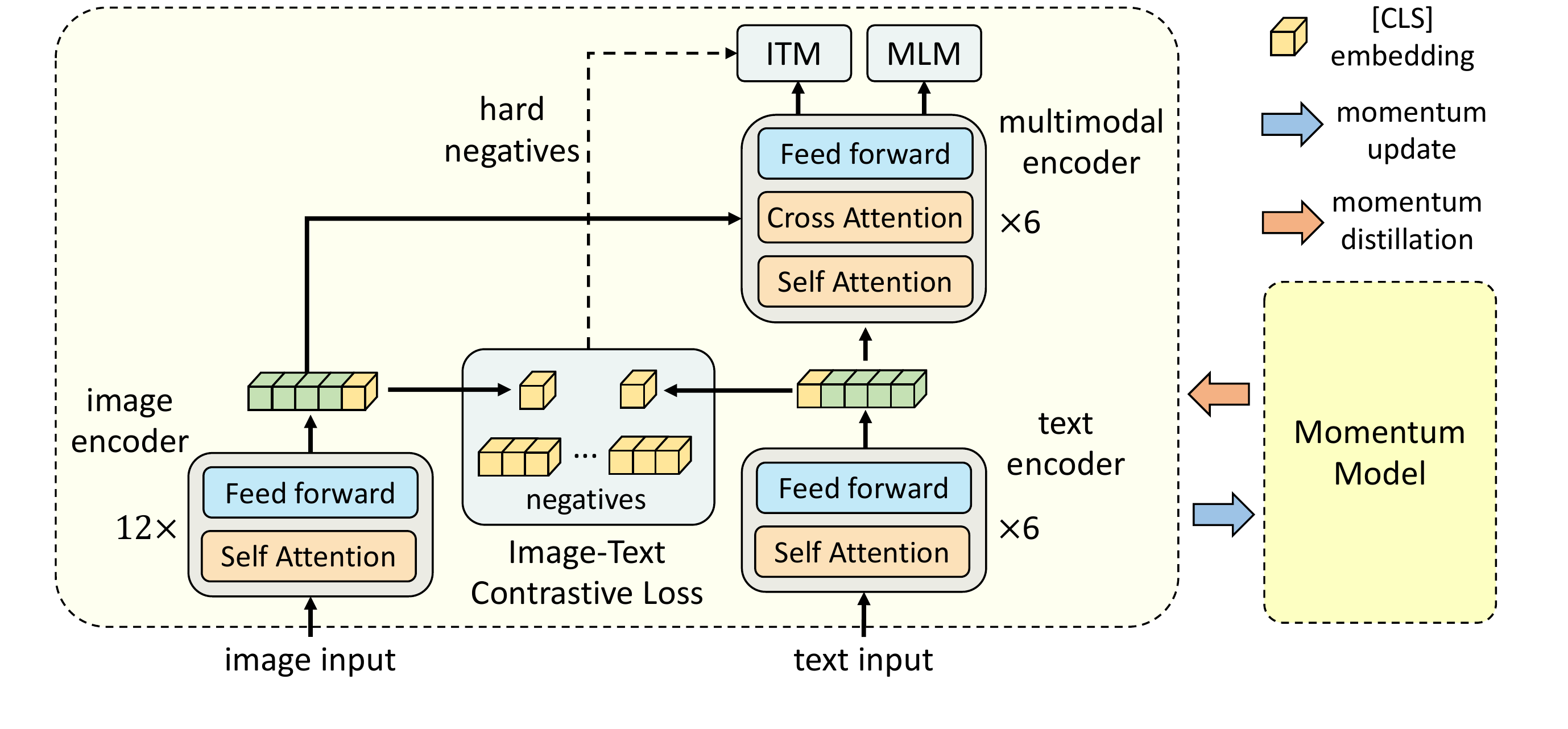}
    \vspace{-4.5ex}
  \caption
  	{ \small
		\textbf{Illustration of ALBEF.} It consists of an image encoder, a text encoder, and a multimodal encoder.
		We propose an image-text contrastive loss to align the unimodal representations of an image-text pair before fusion.
		An image-text matching loss (using in-batch hard negatives mined through contrastive similarity) and a masked-language-modeling loss are applied to learn multimodal interactions between image and text.	
		In order to improve learning with noisy data,
		we generate pseudo-targets using the momentum model (a moving-average version of the base model) as additional supervision during training.
	  } 
  \label{fig:framework}
 \end{figure*} 
In this section,
we first introduce the model architecture (Section~\ref{sec:model}).
Then we delineate the pre-training objectives (Section~\ref{sec:loss}),
followed by the proposed momentum distillation (Section~\ref{sec:mod}).
Lastly we describe the pre-training datasets (Section~\ref{sec:dataset}) and implementation details (Section~\ref{sec:implementation}).

\subsection{Model Architecture}
\label{sec:model}
As illustrated in Figure~\ref{fig:framework},
ALBEF contains an image encoder,
a text encoder, and a multimodal encoder.
We use a 12-layer visual transformer ViT-B/16~\cite{vit} as the image encoder,
and initialize it with weights pre-trained on ImageNet-1k from~\cite{deit}.
An input image $I$ is encoded into a sequence of embeddings:
$\{\Vec{v}_\mathrm{cls}, \Vec{v}_1,...,\Vec{v}_N\}$,
where $v_\mathrm{cls}$ is the embedding of the \texttt{[CLS]} token.
We use a 6-layer transformer~\cite{transformer} for both the text encoder and the multimodal encoder.
The text encoder is initialized using the first 6 layers of the BERT$_\mathrm{base}$~\cite{bert} model,
and the multimodal encoder is initialized using the last 6 layers of the BERT$_\mathrm{base}$. 
The text encoder transforms an input text $T$ into a sequence of embeddings $\{\Vec{w}_\mathrm{cls}, \Vec{w}_1,...,\Vec{w}_N\}$,
which is fed to the multimodal encoder.
The image features are fused with the text features through cross attention at each layer of the multimodal encoder.

\subsection{Pre-training Objectives}
\label{sec:loss}

We pre-train ALBEF with three objectives: image-text contrastive learning (ITC) on the unimodal encoders,
masked language modeling (MLM) and image-text matching (ITM) on the multimodal encoder.
We improve ITM with online contrastive hard negative mining.

\noindent\textbf{Image-Text Contrastive Learning} aims to learn better unimodal representations before fusion.
It learns a similarity function $s=g_v(\Vec{v}_\mathrm{cls})^\top g_w(\Vec{w}_\mathrm{cls})$, such that parallel image-text pairs have higher similarity scores. 
$g_v$ and $g_w$ are linear transformations that map the \texttt{[CLS]} embeddings to normalized lower-dimensional (256-d) representations.
Inspired by MoCo~\cite{moco}, we maintain two queues to store the most recent $M$ image-text representations from the momentum unimodal encoders.
The normalized features from the momentum encoders are denoted as 
$g'_v(\Vec{v}'_\mathrm{cls})$ and $g'_w(\Vec{w}'_\mathrm{cls})$.
We define $s(I,T) = g_v(\Vec{v}_\mathrm{cls})^\top g'_w(\Vec{w}'_\mathrm{cls})$ and $s(T,I) = g_w(\Vec{w}_\mathrm{cls})^\top g'_v(\Vec{v}'_\mathrm{cls})$.

For each image and text,
we calculate the softmax-normalized image-to-text and text-to-image similarity as:
\begin{equation}
\label{eqn:sim}
p_m^\mathrm{i2t}(I) = \frac{\exp (s(I,T_m) / \tau)}{\sum_{m=1}^M \exp (s(I,T_m)/ \tau)}, ~~~~~
p_m^\mathrm{t2i}(T) = \frac{\exp (s(T,I_m)/ \tau)}{\sum_{m=1}^M \exp (s(T,I_m)/ \tau)} 
\end{equation}
where $\tau$ is a learnable temperature parameter. 
Let $\Vec{y}^\mathrm{i2t}(I)$ and $\Vec{y}^\mathrm{t2i}(T)$ denote the ground-truth one-hot similarity,
where negative pairs have a probability of 0 and the positive pair has a probability of 1.
% (or $1/ \text{\#positives}$ in cases of multiple positives).
The image-text contrastive loss is defined as the cross-entropy $\mathrm{H}$ between $\Vec{p}$ and $\Vec{y}$:
\begin{equation}
\label{eqn:itc}
\mathcal{L}_\mathrm{itc} = \frac{1}{2} \mathbb{E}_{(I,T)\sim D} \big[ \mathrm{H}(\Vec{y}^\mathrm{i2t}(I),\Vec{p}^\mathrm{i2t}(I)) + \mathrm{H}(\Vec{y}^\mathrm{t2i}(T),\Vec{p}^\mathrm{t2i}(T)) \big]
\end{equation}

\noindent\textbf{Masked Language Modeling} utilizes both the image and the contextual text to predict the masked words.
We randomly mask out the input tokens with a probability of 15\% and replace them with the special token \texttt{[MASK]}\footnote{following BERT, the replacements are 10\% random tokens, 10\% unchanged, and 80\% \texttt{[MASK]}}.
Let $\hat{T}$ denote a masked text,
and $\Vec{p}^\textrm{msk}(I,\hat{T})$ denote the model's predicted probability for a masked token.
MLM minimizes a cross-entropy loss:
\begin{equation}
\label{eqn:mlm}
\mathcal{L}_\mathrm{mlm} = \mathbb{E}_{(I,\hat{T})\sim D} \mathrm{H} (\Vec{y}^\textrm{msk}, \Vec{p}^\textrm{msk}(I,\hat{T}))
\end{equation}
where $\Vec{y}^\textrm{msk}$ is a one-hot vocabulary distribution where the ground-truth token has a probability of 1.

\noindent\textbf{Image-Text Matching} predicts whether a pair of image and text is positive (matched) or negative (not matched).
We use the multimodal encoder's output embedding of the \texttt{[CLS]} token as the joint representation of the image-text pair, and append a fully-connected (FC) layer followed by softmax to predict a two-class probability $p^\mathrm{itm}$.
The ITM loss is:
\begin{equation}
\label{eqn:itm}
\mathcal{L}_\mathrm{itm} = \mathbb{E}_{(I,T)\sim D} \mathrm{H} (\Vec{y}^\textrm{itm}, \Vec{p}^\textrm{itm}(I,T))
\end{equation}
where $\Vec{y}^\textrm{itm}$ is a 2-dimensional one-hot vector representing the ground-truth label.

We propose a strategy to sample hard negatives for the ITM task with zero computational overhead.
A negative image-text pair is hard if they share similar semantics but differ in fine-grained details.
We use the contrastive similarity from Equation~\ref{eqn:sim} to find in-batch hard negatives.
For each image in a mini-batch,
we sample one negative text from the same batch following the contrastive similarity distribution,
where texts that are more similar to the image have a higher chance to be sampled.
Likewise, we also sample one hard negative image for each text.

The full pre-training objective of ALBEF is:
\begin{equation}
\mathcal{L} = \mathcal{L}_\mathrm{itc}  + \mathcal{L}_\mathrm{mlm} + \mathcal{L}_\mathrm{itm}
\end{equation}

\subsection{Momentum Distillation}

\label{sec:mod}

\begin{figure*}[!t]
	\centering
	\includegraphics[width=\textwidth]{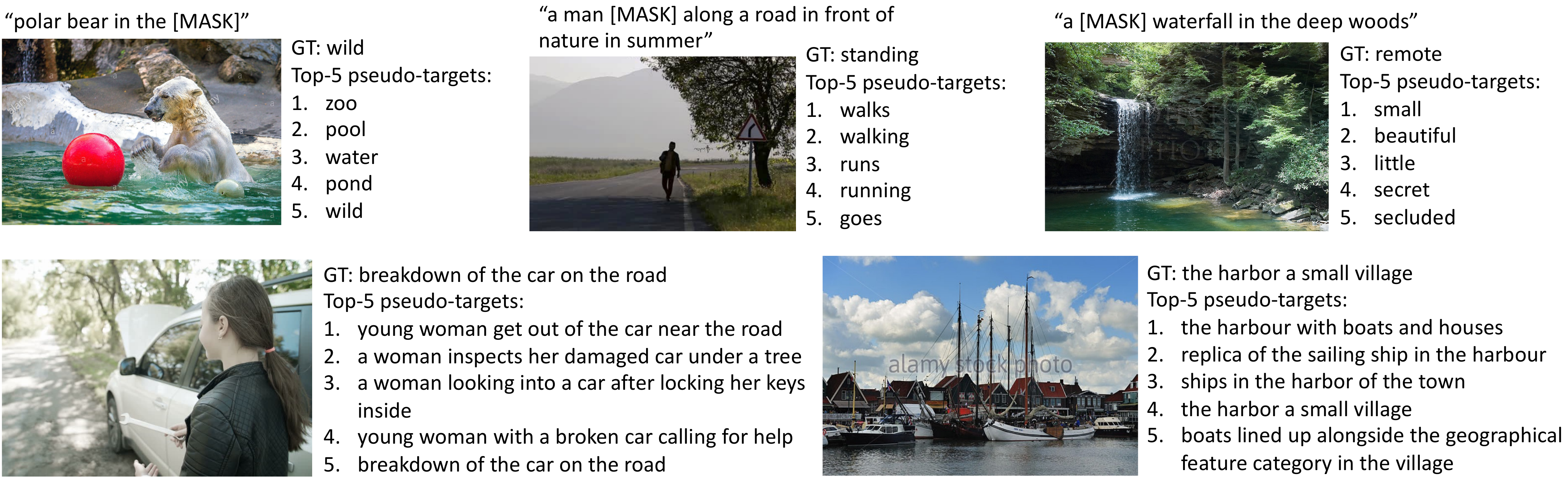}
 \vspace{-3ex}
  \caption
  	{ \small
		Examples of the pseudo-targets for MLM (1st row) and ITC (2nd row).
		The pseudo-targets can capture visual concepts that are not described by the ground-truth text (\eg~``beautiful waterfall'', ``young woman''). 
	  } 
  \label{fig:mod}
 \end{figure*} 

The image-text pairs used for pre-training are mostly collected from the web and they tend to be noisy.
Positive pairs are usually weakly-correlated: the text may contain words that are unrelated to the image,
or the image may contain entities that are not described in the text.
For ITC learning, negative texts for an image may also match the image's content.
For MLM, there may exist other words different from the annotation that describes the image equally well (or better). 
However, the one-hot labels for ITC and MLM penalize all negative predictions regardless of their correctness.

To address this,
we propose to learn from pseudo-targets generated by the momentum model.
The momentum model is a continuously-evolving teacher which
consists of exponential-moving-average versions of the unimodal and multimodal encoders.
During training,
we train the base model such that its predictions match the ones from the momentum model.
Specifically,
for ITC, we first compute the image-text similarity using features from the momentum unimodal encoders as $s'(I,T) = g'_v(\Vec{v}'_\mathrm{cls})^\top g'_w(\Vec{w}'_\mathrm{cls})$ and $s'(T,I) = g'_w(\Vec{w}_\mathrm{cls})^\top g'_v(\Vec{v}'_\mathrm{cls})$. 
Then we compute soft pseudo-targets $\Vec{q}^\mathrm{i2t}$ and $\Vec{q}^\mathrm{t2i}$ by replacing $s$ with $s'$ in Equation~\ref{eqn:sim}.
The ITC$_\mathrm{MoD}$ loss is defined as:
\begin{equation}
\label{eqn:itc_mod}
\mathcal{L}_\mathrm{itc}^\mathrm{mod} = (1-\alpha) \mathcal{L}_\mathrm{itc} + \frac{\alpha }{2} \mathbb{E}_{(I,T)\sim D} \big[ \mathrm{KL}(\Vec{q}^\mathrm{i2t}(I) \parallel \Vec{p}^\mathrm{i2t}(I)) + \mathrm{KL}(\Vec{q}^\mathrm{t2i}(T)\parallel\Vec{p}^\mathrm{t2i}(T))\big]
\end{equation}

Similarly, for MLM, let $\Vec{q}^\textrm{msk}(I,\hat{T})$ denote the momentum model's prediction probability for the masked token,
the MLM$_\mathrm{MoD}$ loss is:
\begin{equation}
\mathcal{L}_\mathrm{mlm}^\mathrm{mod}  = (1-\alpha) \mathcal{L}_\mathrm{mlm} + \alpha \mathbb{E}_{(I,\hat{T})\sim D} \mathrm{KL} (\Vec{q}^\textrm{msk}(I,\hat{T})\parallel \Vec{p}^\textrm{msk}(I,\hat{T}))
\end{equation}
In Figure~\ref{fig:mod},
we show examples of the top-5 candidates from the pseudo-targets, 
which effectively capture relevant words/texts for an image.
More examples can be found in Appendix.

We also apply MoD to the downstream tasks.
The final loss for each task is a weighted combination of the original task's loss and the KL-divergence between the model's prediction and the pseudo-targets.
For simplicity, we set the weight $\alpha=0.4$ for all pre-training and downstream tasks \footnote{our experiments show that $\alpha=0.3,0.4,0.5$ yield similar performance, with $\alpha=0.4$ slightly better}.

\vspace{-0.5ex}
\subsection{Pre-training Datasets}
\vspace{-0.5ex}
\label{sec:dataset}
Following UNITER~\cite{uniter},
we construct our pre-training data using two web datasets (Conceptual Captions~\cite{CC}, SBU Captions~\cite{sbu}) and two in-domain datasets (COCO~\cite{coco} and Visual Genome~\cite{VG}).
The total number of unique images is 4.0M,
and the number of image-text pairs is 5.1M.
To show that our method is scalable with larger-scale web data,
we also include the much noisier Conceptual 12M dataset~\cite{cc12m},
increasing the total number of images to 14.1M~\footnote{some urls provided by the web datasets have become invalid}.
Details are in Appendix.

\vspace{-0.5ex}
\subsection{Implementation Details}
\vspace{-0.5ex}
\label{sec:implementation}
Our model consists of a BERT$_\mathrm{base}$ with 123.7M parameters and a ViT-B/16 with 85.8M parameters.
We pre-train the model for 30 epochs using a batch size of 512 on 8 NVIDIA A100 GPUs.
We use the AdamW~\cite{adamw} optimizer with a weight decay of 0.02.
The learning rate is warmed-up to $1e^{-4}$ in the first 1000 iterations,
and decayed to $1e^{-5}$ following a cosine schedule.
During pre-training,
we take random image crops of resolution $256\times256$ as input,
and also apply RandAugment\footnote{we remove color changes from RandAugment because the text often contains color information}~\cite{randaug}.
During fine-tuning,
we increase the image resolution to $384\times384$ and interpolate the positional encoding of image patches following~\cite{vit}.
The momentum parameter for updating the momentum model is set as 0.995,
and the size of the queue used for image-text contrastive learning is set as 65,536.
We linearly ramp-up the distillation weight $\alpha$ from 0 to 0.4 within the 1st epoch. 
% In the Appendix, we provide additional experimental results using the BERT$_\mathrm{large}$ model.

\section{A Mutual Information Maximization Perspective}
\vspace{-0.5ex}
\label{sec:mutual_info}
In this section,
we provide an alternative perspective of ALBEF and show that it maximizes a lower bound on the mutual information (MI) between different ``views'' of an image-text pair.
ITC, MLM, and MoD can be interpreted as different ways to generate the views.

Formally, we define two random variables $a$ and $b$ as two different views of a data point.
In self-supervised learning~\cite{moco,simclr,goodview}, 
$a$ and $b$ are two augmentations of the same image.
In vision-language representation learning,
we consider $a$ and $b$ as different variations of an image-text pair that capture its semantic meaning.
We aim to learn representations invariant to the change of view.
This can be achieved by maximizing the MI between $a$ and $b$.
In practice, we maximize a lower bound on MI$(a,b)$ by minimizing the InfoNCE loss~\cite{CPC} defined as: 
\begin{equation}
\mathcal{L}_\mathrm{NCE} = -\mathbb{E}_{p(a,b)}\left[ \log \frac{\exp (s(a,b))}{  \sum_{\hat{b}\in\hat{B}} \exp (s(a,\hat{b})) }  \right] 
\end{equation}
% given two random variables $a$ and $b$, mutual information measures their dependency and is defined as 
% \begin{equation}
% 	\mathrm{MI}(a,b)=H(a)-H(a|b)=H(b)-H(b|a)
% \end{equation}

% Since directly maximizing mutual information is intractable,
% recent self-supervised learning methods~\cite{CPC,moco,simclr} propose to maximize a lower bound of MI,
% known as InfoNCE:

% \begin{equation}
% \mathrm{MI}(a,b)\geq \mathbb{E}_{p(a,b)}\left[ \frac{\exp s(a,b)}{  \sum_{\hat{b}\in\hat{B}} \exp s(a,\hat{b}) }  \right] + \log \lvert \hat{B} \rvert
% \end{equation}
where $s(a,b)$ is a scoring function (\eg,~a dot product between two representations), and
$\hat{B}$ contains the positive sample $b$ and $\lvert \hat{B} \rvert-1$ negative samples drawn from a proposal distribution.

Our ITC loss with one-hot labels (Equation~\ref{eqn:itc}) can be re-written as:
\begin{equation}
\mathcal{L}_\mathrm{itc} = -\frac{1}{2} \mathbb{E}_{p(I,T)} \big[\log \frac{\exp (s(I,T) / \tau)}{\sum_{m=1}^M \exp (s(I,T_m)/ \tau)} + \log \frac{\exp (s(T,I) / \tau)}{\sum_{m=1}^M \exp (s(T,I_m)/ \tau)}\big]
\end{equation}

Minimizing $\mathcal{L}_\mathrm{itc}$ can be seen as maximizing a symmetric version of InfoNCE.
Hence, ITC considers the two individual modalities (\ie,~$I$ and $T$) as the two views of an image-text pair,  
and trains the unimodal encoders to maximize the MI between the image and text views for the positive pairs.

As shown in~\cite{MI}, we can also interpret MLM as maximizing the MI between a masked word token and its masked context (\ie~image + masked text).
Specifically, we can re-write the MLM loss with one-hot labels (Equation~\ref{eqn:mlm}) as
\begin{equation}
\mathcal{L}_\mathrm{mlm} = - \mathbb{E}_{p(I,\hat{T})} \big[\log \frac{\exp  
	(\psi(y^\mathrm{msk})^\top  f(I,\hat{T}))} { \sum_{y \in \mathcal{V}} \exp (\psi(y)^\top  f(I,\hat{T}))} \big]
\end{equation}
where $\psi(y): \mathcal{V} \rightarrow \mathbb{R}^d$  is a lookup function in the multimodal encoder's output layer that maps a word token $y$ into a vector and $\mathcal{V}$ is the full vocabulary set,
and $f(I,\hat{T})$ is a function that returns the final hidden state of the multimodal encoder corresponding to the masked context.
Hence, MLM considers the two views of an image-text pair to be: (1) a randomly selected word token,
and (2) the image + the contextual text with that word masked.

Both ITC and MLM generate views by taking partial information from an image-text pair, through either modality separation or word masking.
Our momentum distillation can be considered as generating alternative views from the entire proposal distribution.
Take ITC$_\mathrm{MoD}$ in Equation~\ref{eqn:itc_mod} as an example,
minimizing {\small$\mathrm{KL}(\Vec{p}^\mathrm{i2t}(I),\Vec{q}^\mathrm{i2t}(I))$} is equivalent to minimizing the following objective:
\begin{equation}
 -\sum_m  {q}_m^\mathrm{i2t}(I) \log {p}_m^\mathrm{i2t}(I)
 = -\sum_m  \frac{\exp (s'(I,T_m) / \tau)}{\sum_{m=1}^M \exp (s'(I,T_m)/ \tau)} 
\log \frac{\exp (s(I,T_m) / \tau)}{\sum_{m=1}^M \exp (s(I,T_m)/ \tau)}
\end{equation}
It maximizes MI$(I,T_m)$ for texts that share similar semantic meaning with the image $I$ because those texts would have larger ${q}_m^\mathrm{i2t}(I)$.
Similarly,
ITC$_\mathrm{MoD}$ also maximizes MI$(I_m,T)$ for images that are similar to $T$.
We can follow the same method to show that MLM$_\mathrm{MoD}$ generates alternative views $y'\in \mathcal{V}$ for the masked word $y^\mathrm{msk}$,
and maximizes the MI between $y'$ and $(I,\hat{T})$.
Therefore, our momentum distillation can be considered as performing data augmentation to the original views.
The momentum model generates a diverse set of views that are absent in the original image-text pairs,
and encourages the base model to learn representations that capture view-invariant semantic information.

\vspace{-0.5ex}
\section{Downstream V+L Tasks}
\vspace{-0.5ex}
We adapt the pre-trained model to five downstream V+L tasks.
We introduce each task and our fine-tuning strategy below. 
Details of the datasets and fine-tuning hyperparameters are in Appendix.

\noindent \textbf{Image-Text Retrieval} contains two subtasks:
image-to-text retrieval (TR) and text-to-image retrieval (IR).
We evaluate ALBEF on the Flickr30K~\cite{flickr} and COCO benchmarks,
and fine-tune the pre-trained model using the training samples from each dataset.
For zero-shot retrieval on Flickr30K, we evaluate with the model fine-tuned on COCO.
During fine-tuning, we jointly optimize the ITC loss (Equation~\ref{eqn:itc}) and the ITM loss (Equation~\ref{eqn:itm}).
ITC learns an image-text scoring function based on similarity of unimodal features,
whereas ITM models the fine-grained interaction between image and text to predict a matching score.
Since the downstream datasets contain multiple texts for each image,
we change the ground-truth label of ITC to consider multiple positives in the queue, where each positive has a ground-truth probability of $1/\#\mathrm{positives}$.
During inference,
we first compute the feature similarity score $s_\mathrm{itc}$ for all image-text pairs.
Then we take the top-$k$ candidates and calculate their ITM score $s_\mathrm{itm}$ for ranking.
%The final score is computed as $s_\mathrm{itc} / \tau_\mathrm{test} + s_\mathrm{itm}$,
%where $\tau_\mathrm{test}$ is a scaling hyperparameter determined by validation performance.
Because $k$ can be set to be very small,
our inference speed is much faster than methods that require computing the ITM score for all image-text pairs~\cite{uniter,oscar,villa}.

\noindent \textbf{Visual Entailment} (SNLI-VE\footnote{results on SNLI-VE should be interpreted with caution because its test data has been reported to be noisy~\cite{ve-2.0}}~\cite{VE}) is a fine-grained visual reasoning
task to predict whether the relationship between an image and a text is entailment, neutral, or contradictory.
We follow UNITER~\cite{uniter} and consider VE as a three-way classification problem,
and predict the class probabilities using a multi-layer perceptron (MLP) on the multimodal encoder's representation of the \texttt{[CLS]} token.

\begin{figure*}[!t]
  \vspace{-2ex}
	\centering
	\includegraphics[width=\textwidth]{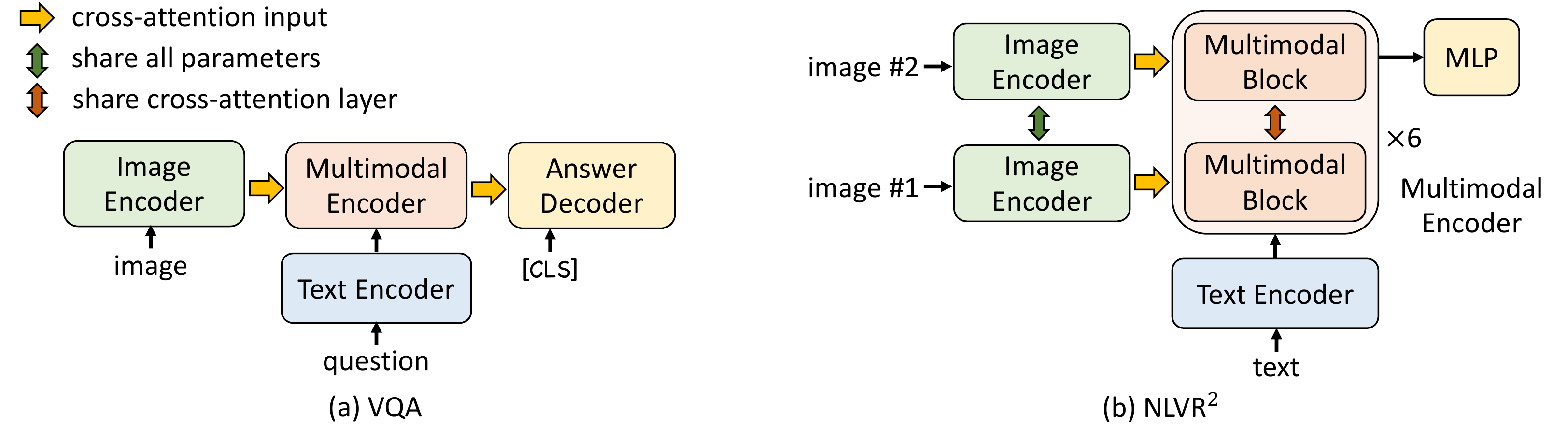}
	\vspace{-3ex}
  \caption
  	{ \small
		The model architecture for VQA and NLVR$^2$. 
		For VQA, we append an auto-regressive decoder to generate the answer given the image-question embeddings.
		For NLVR$^2$, we replicate the transformer block within each layer of multimodal encoder to enable reasoning over two images.
	  } 
  \label{fig:vqa_nlvr}
  \vspace{-2.5ex}
 \end{figure*}

\noindent \textbf{Visual Question Answering} (VQA~\cite{VQA2}) requires the model to predict an answer given an image and a question.
Different from existing methods that formulate VQA as a multi-answer classification problem~\cite{MAttNet,uniter},
we consider VQA as an answer generation problem, similar to~\cite{VL_T5}.
Specifically, we use a 6-layer transformer decoder to generate the answer.
As shown in Figure~\ref{fig:vqa_nlvr}a, 
the auto-regressive answer decoder receives the multimodal embeddings through cross attention,
and a start-of-sequence token (\texttt{[CLS]}) is used as the decoder's initial input token. 
Likewise, an end-of-sequence token (\texttt{[SEP]}) is appended to the end of decoder
outputs which indicates the completion of generation.
The answer decoder is initialized using the pre-trained weights from the multimodal encoder,
and finetuned with a conditional language-modeling loss.
For a fair comparison with existing methods,
we constrain the decoder to only generate from the 3,192 candidate answers~\cite{bilinear} during inference.

\noindent \textbf{Natural Language for Visual Reasoning} (NLVR$^2$~\cite{NLVR})
requires the model to predict whether a text describes a pair of images.
We extend our multimodal encoder to enable reasoning over two images.
As shown in Figure~\ref{fig:vqa_nlvr}b,
each layer of the multimodal encoder is replicated to have two consecutive transformer blocks, 
where each block contains a self-attention layer, a cross-attention layer, and a feed-forward layer (see Figure~\ref{fig:framework}).
The two blocks within each layer are initialized using the same pre-trained weights,
and the two cross-attention layers share the same linear projection weights for the keys and values.
During training,
the two blocks receive two sets of image embeddings for the image pair.
We append a MLP classifier on the multimodal encoder's \texttt{[CLS]} representation for prediction.
% Following UNITER~\cite{uniter}, we learn a MLP classifier on the multimodal encoder's representation of the \texttt{[CLS]} token to predict ``true'' or ``false''.

For NLVR$^2$, we perform an additional pre-training step to prepare the new multimodal encoder for encoding an image-pair.
We design a text-assignment (TA) task as follows:
given a pair of images and a text, the model needs to assign the text to either the first image, the second image, or none of them.
We consider it as a three-way classification problem, and use a FC layer on the \texttt{[CLS]} representation to predict the assignment.
We pre-train with TA for only 1 epoch using the 4M images (Section~\ref{sec:dataset}).

\noindent \textbf{Visual Grounding} 
aims to localize the region in an image that corresponds to a specific textual description.
We study the weakly-supervised setting,
where no bounding box annotations are available.
We perform experiments on the RefCOCO+~\cite{refer} dataset,
and fine-tune the model using only image-text supervision following the same strategy as image-text retrieval.
During inference,
we extend Grad-CAM~\cite{gradcam} to acquire heatmaps, and use them to rank the detected proposals provided by~\cite{MAttNet}.

   \vspace{-1ex}
\section{Experiments}
\vspace{-0.5ex}
\label{sec:experiment}

\begin{table*}[!b]
    \small
    \vspace{-1ex}
    \aboverulesep = 0.55mm
    \belowrulesep = 0.55mm
	\centering	
  % \setlength\tabcolsep{5pt}
%	\resizebox{\textwidth}{!}{%
	\begin{tabular}	{l  l |  c  c  c  c  c  }
		\toprule	 	
	 \#Pre-train & \multirow{2}{*}{Training tasks} & TR &  IR & SNLI-VE & NLVR$^2$ & VQA\\
	 Images &  & \multicolumn{2}{c}{(flickr test)} & (test) & (test-P) & (test-dev) \\
	 	 \midrule
	 \multirow{6}{*}{4M}	 & MLM + ITM & 93.96&88.55 & 77.06 &77.51 & 71.40 \\
	 	 & ITC + MLM + ITM & 96.55 & 91.69 & 79.15& 79.88& 73.29\\ 
	 	 & ITC + MLM + ITM$_\text{hard}$  & 97.01& 92.16 & 79.77&80.35 & 73.81\\ 
	 	 & ITC$_\text{MoD}$ + MLM + ITM$_\text{hard}$ &97.33 & 92.43 &79.99 & 80.34& 74.06\\ 
	 	 & Full  (ITC$_\text{MoD}$ + MLM$_\text{MoD}$ + ITM$_\text{hard}$) & 97.47 & 92.58 & 80.12 &80.44 &74.42 \\ 
	 	 & ALBEF (Full + MoD$_\text{Downstream}$) & 97.83 & 92.65 & 80.30 & 80.50 & 74.54 \\ 	 
	 \midrule
	 14M	 & ALBEF  & 98.70&94.07 & 80.91 & 83.14 & 75.84 \\

		\bottomrule
	\end{tabular}
%	}
	\caption
	{
	\small	
		Evaluation of the proposed methods on four downstream V+L tasks.
		For text-retrieval (TR) and image-retrieval (IR), we report the average of R@1, R@5 and R@10.
		ITC: image-text contrastive learning.
		MLM: masked language modeling.
		ITM$_\text{hard}$: image-text matching with contrastive hard negative mining.
		MoD: momentum distillation.		
		MoD$_\text{Downstream}$: momentum distillation on downstream tasks.
	}
	\label{tbl:main}
	\vspace{-2ex}

\end{table*}		
\vspace{-0.5ex}
\subsection{Evaluation on the Proposed Methods}
\vspace{-0.5ex}
First,
we evaluate the effectiveness of the proposed methods (\ie~image-text contrastive learning, contrastive hard negative mining, and momentum distillation).
Table~\ref{tbl:main} shows the performance of the downstream tasks with different variants of our method.
Compared to the baseline pre-training tasks (MLM+ITM),
adding ITC substantially improves the pre-trained model's performance across all tasks.
The proposed hard negative mining improves ITM by finding more informative training samples.
Furthermore,
adding momentum distillation improves learning for both ITC (row 4), MLM (row 5), and on all downstream tasks (row 6).
In the last row, we show that ALBEF can effectively leverage more noisy web data to improve the pre-training performance.
\begin{table*}[!t]
    \aboverulesep = 0.48mm
    \belowrulesep = 0.48mm
    \small
	\centering	
 \setlength\tabcolsep{4pt}
 \vspace{-3ex}
	\resizebox{\textwidth}{!}{%
	\begin{tabular}	{l  c |  c  c  c  c  c  c | c  c  c  c  c  c }
		\toprule	 	
	 \multirow{2}{*}{Method} & \# Pre-train & \multicolumn{6}{c|}{Flickr30K (1K test set)} & \multicolumn{6}{c}{MSCOCO (5K test set)} \\
	 & ~~Images &  \multicolumn{3}{c}{TR}& \multicolumn{3}{c|}{IR} &  \multicolumn{3}{c}{TR}& \multicolumn{3}{c}{IR}\\
	 \midrule
	& & R@1 &R@5&R@10& R@1 &R@5&R@10& R@1 &R@5&R@10& R@1 &R@5&R@10\\
	UNITER & 4M & 87.3 & 98.0 &99.2 &75.6 &94.1 &96.8 &65.7 &88.6 &93.8 &52.9 &79.9 &88.0 \\
	VILLA & 4M &87.9 & 97.5 & 98.8 & 76.3 & 94.2 & 96.8 & - & - & - & - & - & - \\
	OSCAR & 4M  & - & - & - & - & - & - & 70.0 & 91.1 & 95.5 &  54.0 & 80.8 & 88.5\\
	ALIGN & 1.2B& 95.3 &  99.8 & 100.0 & 84.9 & 97.4 & 98.6 & 77.0 & 93.5  & 96.9 & 59.9 & 83.3 & 89.8\\
	\midrule
	ALBEF & 4M & 94.3 & 99.4 & 99.8 & 82.8 & 96.7 & 98.4 & 73.1 &91.4 &96.0 &56.8 &81.5 &89.2 \\
	ALBEF & 14M & \textbf{95.9} & \textbf{99.8}  & \textbf{100.0}  &   \textbf{85.6}& \textbf{97.5}  &\textbf{98.9} &
	\textbf{77.6} & \textbf{94.3}& \textbf{97.2} & \textbf{60.7} & \textbf{84.3} & \textbf{90.5} \\
		\bottomrule
	\end{tabular}
	}
	\vspace{-0.8ex}
	\caption
	{
	\small	
		Fine-tuned image-text retrieval results on Flickr30K and COCO datasets.
	}
	\vspace{-0.7ex}
	\label{tbl:retrieval}

\end{table*}		
\vspace{-0.7ex}
\subsection{Evaluation on Image-Text Retrieval}
\vspace{-0.8ex}

\begin{table*}[!t]
    \small
	\centering	
    \aboverulesep = 0.46mm
    \belowrulesep = 0.46mm
	\resizebox{0.75\textwidth}{!}{%
	\begin{tabular}	{l  c |  c  c  c  c  c  c  }
		\toprule	 	
	 \multirow{2}{*}{Method} & \# Pre-train & \multicolumn{6}{c}{Flickr30K (1K test set)}\\
	 & ~~Images &  \multicolumn{3}{c}{TR}& \multicolumn{3}{c}{IR} \\
	 \midrule
	& & R@1 &R@5&R@10& R@1 &R@5&R@10 \\
	%ImageBERT~\cite{ImageBERT} & 2M &70.7 & 90.2 & 94.0 & 54.3 & 79.6 & 87.5 \\
	UNITER~\cite{uniter} & 4M & 83.6 & 95.7 & 97.7 & 68.7 & 89.2 & 93.9 \\
	CLIP~\cite{clip} & 400M & 88.0 & 98.7 & 99.4 & 68.7 & 90.6 & 95.2 \\
	ALIGN~\cite{align} & 1.2B& 88.6 &98.7& 99.7& 75.7& 93.8& 96.8 \\
	\midrule
	ALBEF & 4M & 90.5 & 98.8 & 99.7 & 76.8&93.7 &96.7 \\
	ALBEF & 14M & \textbf{94.1} & \textbf{99.5}& \textbf{99.7} &\textbf{82.8} & \textbf{96.3} &  \textbf{98.1} \\
		\bottomrule
	\end{tabular}
 	}
 	 \vspace{-0.8ex}
	\caption
	{
	\small	
		Zero-shot image-text retrieval results on Flickr30K.
	}
	\label{tbl:retrieval_zeroshot}
    \vspace{-0.7ex}
\end{table*}

Table~\ref{tbl:retrieval} and Table~\ref{tbl:retrieval_zeroshot} report results on fine-tuned and zero-shot image-text retrieval,
respectively.
Our ALBEF achieves state-of-the-art performance,
outperforming CLIP~\cite{clip} and ALIGN~\cite{align} which are trained on orders of magnitude larger datasets.
Given the considerable amount of improvement of ALBEF when the number of training images increases from 4M to 14M,
we hypothesize that it has potential to further grow by training on larger-scale web image-text pairs.

\begin{table*}[!t]
    \aboverulesep = 0.48mm
    \belowrulesep = 0.48mm
    \small
	\centering	
	\resizebox{0.72\textwidth}{!}{%
	\begin{tabular}	{l   |  c  c  c  c  c  c  }
		\toprule	 	
	 \multirow{2}{*}{Method} & \multicolumn{2}{c}{VQA} & \multicolumn{2}{c}{NLVR$^2$} & \multicolumn{2}{c}{SNLI-VE} \\
	  & test-dev & test-std & dev & test-P & val & test\\
	  \midrule
	  VisualBERT~\cite{VisualBERT} & 70.80 & 71.00 & 67.40 & 67.00 & - & - \\
	  VL-BERT~\cite{VL-BERT} & 71.16 & - &  - & - & - & - \\
	  LXMERT~\cite{LXMERT}  & 72.42 & 72.54 & 74.90 & 74.50 & - & - \\
	  12-in-1~\cite{12in1} & 73.15 & - & - & 78.87 & - & 76.95 \\
	  UNITER~\cite{uniter} & 72.70 & 72.91 & 77.18 & 77.85 & 78.59 & 78.28 \\
	   VL-BART/T5~\cite{VL_T5} & - & 71.3 & - & 73.6& - & - \\
	   ViLT~\cite{ViLT}  & 70.94 & - & 75.24 & 76.21& - & - \\
	   OSCAR~\cite{oscar} &  73.16 & 73.44 & 78.07 & 78.36 & - & - \\
	   VILLA~\cite{villa} & 73.59 & 73.67 & 78.39 & 79.30 & 79.47 & 79.03 \\
	  \midrule
	  ALBEF (4M) & 74.54 & 74.70 & 80.24 & 80.50 & 80.14 & 80.30 \\
	   ALBEF (14M) & \textbf{75.84} & \textbf{76.04} & \textbf{82.55} &  \textbf{83.14}& \textbf{80.80} &  \textbf{80.91} \\
		\bottomrule
	\end{tabular}
 	}
 \vspace{-0.8ex}
	\caption
	{
	\small	
		Comparison with state-of-the-art methods on downstream vision-language tasks.
	}
	\label{tbl:results}
	\vspace{-3.5ex}

\end{table*}		 
\vspace{-1.2ex}
\subsection{Evaluation on VQA, NLVR, and VE}
\vspace{-0.8ex}
Table~\ref{tbl:results} reports the comparison with existing methods on other V+L understanding tasks.
With 4M pre-training images,
ALBEF already achieves state-of-the-art performance.
With 14M pre-training images,
ALBEF substantially outperforms existing methods,
including methods that additionally use object tags~\cite{oscar}
or adversarial data augmentation~\cite{villa}. 
Compared to VILLA~\cite{villa},
ALBEF achieves absolute improvements of $2.37\%$ on VQA test-std, 
$3.84\%$ on NLVR$^2$ test-P,
and $1.88\%$ on SNLI-VE test.
Because ALBEF is detector-free and requires lower resolution images,
it also enjoys much faster inference speed compared to most existing methods (>10 times faster than VILLA on NLVR$^2$).

\vspace{-1.2ex}
\subsection{Weakly-supervised Visual Grounding}
\vspace{-0.8ex}

Table~\ref{tbl:grounding} shows the results on RefCOCO+,
where ALBEF substantially outperforms existing methods~\cite{ARN,CCL} (which use weaker text embeddings). 
The ALBEF$_\mathrm{itc}$ variant computes Grad-CAM visualizations on the self-attention maps in the last layer of the image encoder, 
where the gradients are acquired by maximizing the image-text similarity $s_\mathrm{itc}$.
The ALBEF$_\mathrm{itm}$ variant computes Grad-CAM on the cross-attention maps in the 3rd layer of the multimodal encoder (which is a layer specialized in grounding),
where the gradients are acquired by maximizing the image-text matching score $s_\mathrm{itm}$.
Figure~\ref{fig:refcoco} provides a few visualizations. More analysis is in Appendix.

\begin{figure*}[!t]
	\centering
	\includegraphics[width=\textwidth]{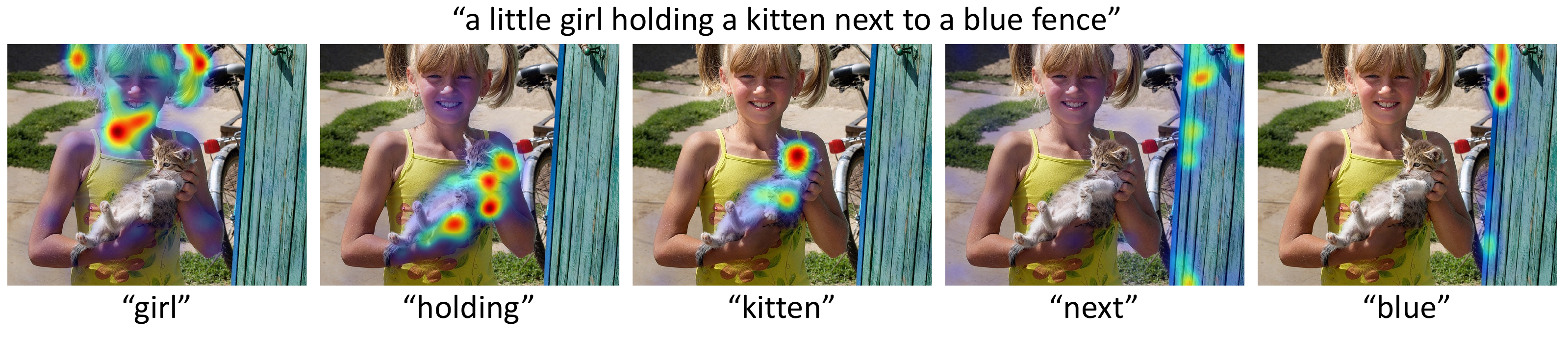}
 \vspace{-4ex}
  \caption
  	{ \small
		Grad-CAM visualizations on the cross-attention maps corresponding to individual words.
	  } 
  \label{fig:coco}
   \vspace{-3ex}
 \end{figure*}

We provide the Grad-CAM visualizations for VQA in Figure~\ref{fig:vqa}.
As can be seen in Appendix, the Grad-CAM visualizations from ALBEF are highly correlated with where humans would look when making decisions.
In Figure~\ref{fig:coco}, we show per-word visualizations for COCO.
Notice how our model not only grounds objects, but also their attributes and relationships.

\begin{table}[!t]
\vspace{-4ex}
\begin{minipage}[b]{0.37\textwidth}
    \aboverulesep = 0.55mm
    \belowrulesep = 0.55mm
	\small
	\centering	
     \setlength\tabcolsep{4pt}
		\begin{tabular}	{l  |  c c c }
			\toprule
			Method & Val & TestA & TestB \\
% 			\midrule
% 			\multirow{2}{*}{ZS} & CLIP~\cite{clip} & & & \\
% 			& ALBEF & & & \\
			\midrule
			ARN~\cite{ARN} & 32.78 & 34.35 & 32.13 \\
			CCL~\cite{CCL} & 34.29& 36.91 & 33.56 \\
			ALBEF$_\mathrm{itc}$ & 51.58 & 60.09 & 40.19\\
			ALBEF$_\mathrm{itm}$ & \textbf{58.46} & \textbf{65.89} & \textbf{46.25}\\			
			\bottomrule
		\end{tabular}
		\vspace{1ex}
    \caption
	{
		\small	
		Weakly-supervised visual \\ grounding on RefCOCO+~\cite{refer} dataset.
	}
	\label{tbl:grounding}	
\end{minipage}
\hfill
\begin{minipage}[b]{0.61\textwidth}
%\vspace{-0.5ex}
	\centering
	\includegraphics[width=\textwidth]{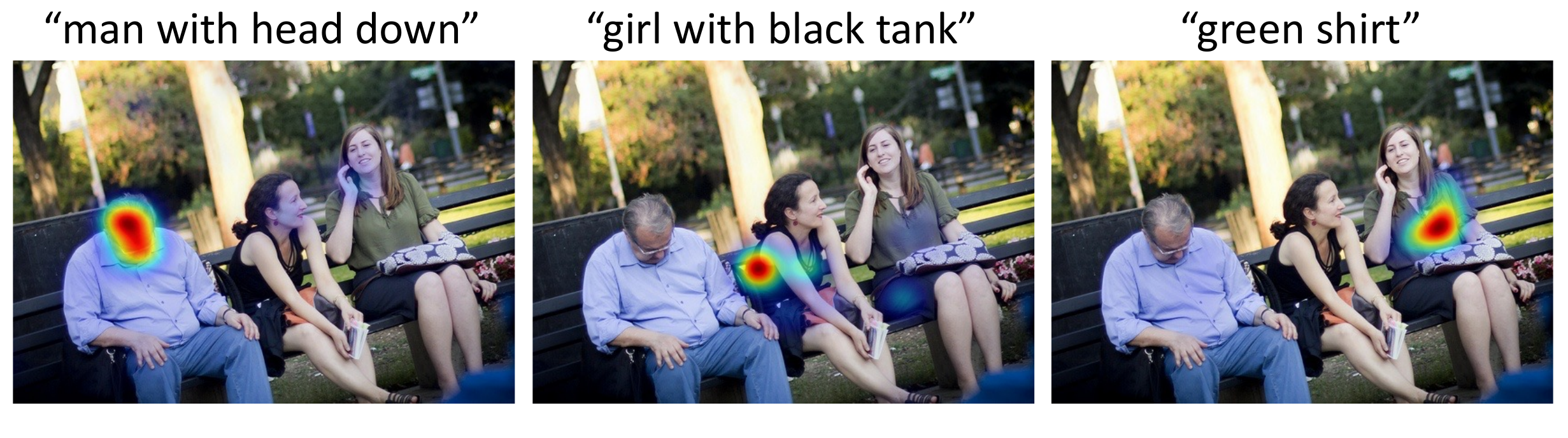}
	\vspace{-3ex}
	    \captionof{figure}
	{
		\small	
		Grad-CAM visualization on the cross-attention maps in the 3rd layer of the multimodal encoder.
	}
	\label{fig:refcoco}
\end{minipage}
\vspace{-3.5ex}
\end{table}
\begin{figure*}[!t]
	\centering
	\includegraphics[width=\textwidth]{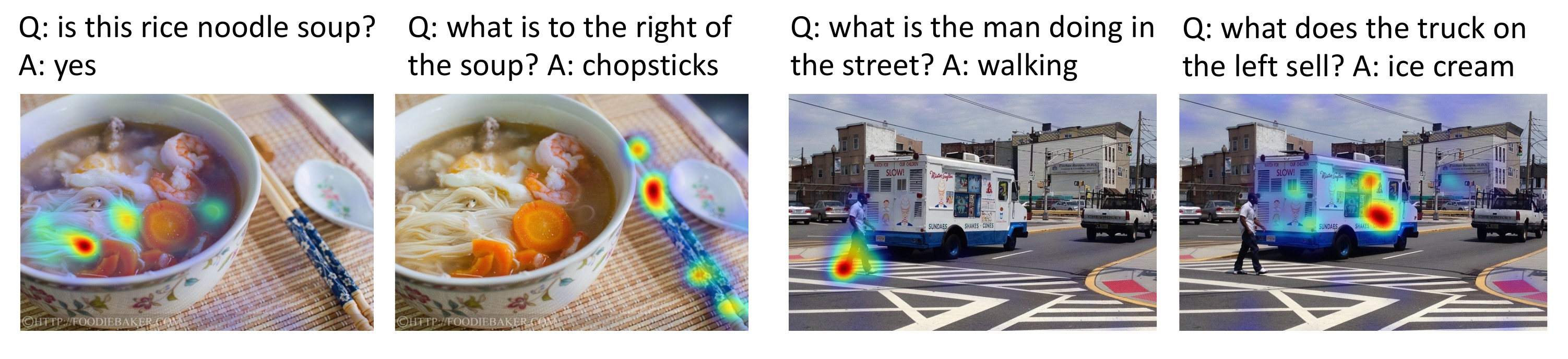}
 \vspace{-5ex}
  \caption
  	{ \small
		Grad-CAM visualizations on the cross-attention maps of the multimodal encoder for the VQA model.
	  } 
  \label{fig:vqa}
   \vspace{-2.5ex}
 \end{figure*} 

\vspace{-1.3ex}
\subsection{Ablation Study}
\vspace{-1ex}

\begin{wraptable}{r}{8.8cm}
	    \aboverulesep = 0.48mm
    \belowrulesep = 0.48mm
    \vspace{-5ex}
	\small
	\setlength\tabcolsep{4pt}
	\centering	
	\begin{tabular}	{c | c c c c  | c }
		\toprule
		\multirow{2}{*}{Flickr30K} & \multicolumn{4}{c|}{w/ hard negs} &  w/o hard negs\\
		&  $s_\mathrm{itc}$& $k=16$  & $k=128$ & $k=256$&  $k=128$ \\
		\midrule
		TR & 97.30 &98.60 &98.57& 98.57 & 98.22 ($-$0.35)  \\
		IR & 90.95 & 93.64  &93.99 &93.95 & 93.68 ($-$0.31)\\
		\bottomrule
	\end{tabular}
	%	}
	\vspace{-0.8ex}
	\caption
	{
		\small	
		Ablation study on fine-tuned image-text retrieval. The average recall on the test set is reported. We use $s_\mathrm{itc}$ to filter top-$k$ candidates and calculate their $s_\mathrm{itm}$ score for ranking. 
	}
	\label{tbl:retrieval_ablation}
	\vspace{-2ex}
\end{wraptable} 

% \begin{table*}[!t]
% 	\small
% 	\centering	
% 	\begin{tabular}	{c | c c c c c  | c }
% 		\toprule
% 		\multirow{2}{*}{Flickr30K} & \multicolumn{5}{c|}{w/ hard negs} &  w/o hard negs\\
% 		&  $s_\mathrm{itc}$& $k=16$ & $k=32$  & $k=128$ & $k=256$&  $k=128$ \\
% 		\midrule
% 		TR & 97.30 &98.60 & 98.57 &98.57& 98.57 & 98.22 ($-$0.35)  \\
% 		IR & 90.95 & 93.64 & 93.54 &93.99 &93.95 & 93.68 ($-$0.31)\\
% 		\bottomrule
% 	\end{tabular}
% 	%	}
% 	\caption
% 	{
% 		\small	
% 		Ablation study on fine-tuned image-text retrieval. The average recall on the test set is reported. We use $s_\mathrm{itc}$ to filter top-$k$ candidates and calculate their $s_\mathrm{itm}$ score for ranking. 
% 	}
% 	\label{tbl:retrieval_ablation}
	
% \end{table*}					
Table~\ref{tbl:retrieval_ablation} studies the effect of various design choices on image-text retrieval.
Since we use $s_\mathrm{itc}$ to filter top-$k$ candidates during inference,
we vary $k$ and report its effect.
In general, the ranking result acquired by $s_\mathrm{itm}$ is not sensitive to changes in $k$.
% The reason is that using $s_\mathrm{itc}$ alone can already achieve good recalls, hence the top-$k$ candidates most likely contain the correct one.
We also validate the effect of hard negative mining in the last column.

\begin{wraptable}{r}{8.8cm}
	    \aboverulesep = 0.4mm
    \belowrulesep = 0.4mm
	\small
	\centering	
	\vspace{-2.5ex}
	\setlength\tabcolsep{2pt}
\begin{tabular}	{c | c c c | c c  c}
	\toprule
	\multirow{2}{*}{NLVR$^2$} & \multicolumn{3}{c|}{w/ TA} &  \multicolumn{3}{c}{w/o TA}\\
	 & share all & share CA  & no share & share all &  share CA & no share \\
	\midrule
	dev &  82.13& 82.55 & 81.93 & 80.52 & 80.28 & 77.84\\
	test-P &   82.36 & 83.14 & 82.85 & 81.29 & 80.45 & 77.58\\
		\bottomrule
\end{tabular}
%	}
\vspace{-1ex}
\caption
{
\small	
Ablation study on NLVR$^2$.
}
\label{tbl:nlvr_ablation}
	\vspace{-2.5ex}
\end{wraptable} 		
Table~\ref{tbl:nlvr_ablation} studies the effect of text-assignment (TA) pre-training and parameter sharing on NLVR$^2$.
We examine three strategies: (1) the two mutimodal blocks share all parameters, (2) only the cross-attention (CA) layers are shared, (3) no sharing.
Without TA, sharing the entire block has better performance.
With TA to pre-train the model for image-pair, sharing CA leads to the best performance.

   \vspace{-1.2ex}
\section{Conclusion and Social Impacts}
\vspace{-1.2ex}
\label{sec:conclusion}
This paper proposes ALBEF, a new framework for vision-language representation learning.
ALBEF first aligns the unimodal image representation and text representation before fusing them with a multimodal encoder.
We theoretically and experimentally verify the effectiveness of the proposed image-text contrastive learning and momentum distillation.
Compared to existing methods, ALBEF offers better performance and faster inference speed on multiple downstream V+L tasks.

While our paper shows promising results on vision-language representation learning,
additional analysis on the data and the model is necessary before deploying it in practice,
because
web data may contain unintended private information, unsuitable images, or harmful texts,
and only optimizing accuracy may have unwanted social implications.

   % \vspace{-0.5ex}
% \section{Social Impacts and Future Work}
% \vspace{-0.5ex}
% \label{sec:social}

% % The unwanted bias is also prone to be amplified by machine learning models.

\bibliographystyle{nips}
\bibliography{bib}

\begin{thebibliography}{10}

\bibitem{LXMERT}
Tan, H., M.~Bansal.
\newblock {LXMERT:} learning cross-modality encoder representations from
  transformers.
\newblock In K.~Inui, J.~Jiang, V.~Ng, X.~Wan, eds., \emph{{EMNLP}}, pages
  5099--5110. 2019.

\bibitem{uniter}
Chen, Y., L.~Li, L.~Yu, et~al.
\newblock {UNITER:} universal image-text representation learning.
\newblock In \emph{ECCV}, vol. 12375, pages 104--120. 2020.

\bibitem{oscar}
Li, X., X.~Yin, C.~Li, et~al.
\newblock Oscar: Object-semantics aligned pre-training for vision-language
  tasks.
\newblock In \emph{{ECCV}}, pages 121--137. 2020.

\bibitem{CC}
Sharma, P., N.~Ding, S.~Goodman, et~al.
\newblock Conceptual captions: {A} cleaned, hypernymed, image alt-text dataset
  for automatic image captioning.
\newblock In I.~Gurevych, Y.~Miyao, eds., \emph{{ACL}}, pages 2556--2565. 2018.

\bibitem{sbu}
Ordonez, V., G.~Kulkarni, T.~L. Berg.
\newblock Im2text: Describing images using 1 million captioned photographs.
\newblock In J.~Shawe{-}Taylor, R.~S. Zemel, P.~L. Bartlett, F.~C.~N. Pereira,
  K.~Q. Weinberger, eds., \emph{{NIPS}}, pages 1143--1151. 2011.

\bibitem{clip}
Radford, A., J.~W. Kim, C.~Hallacy, et~al.
\newblock Learning transferable visual models from natural language
  supervision.
\newblock \emph{arXiv preprint arXiv:2103.00020}, 2021.

\bibitem{align}
Jia, C., Y.~Yang, Y.~Xia, et~al.
\newblock Scaling up visual and vision-language representation learning with
  noisy text supervision.
\newblock \emph{arXiv preprint arXiv:2102.05918}, 2021.

\bibitem{villa}
Gan, Z., Y.~Chen, L.~Li, et~al.
\newblock Large-scale adversarial training for vision-and-language
  representation learning.
\newblock In H.~Larochelle, M.~Ranzato, R.~Hadsell, M.~Balcan, H.~Lin, eds.,
  \emph{{NeurIPS}}. 2020.

\bibitem{gradcam}
Selvaraju, R.~R., M.~Cogswell, A.~Das, et~al.
\newblock Grad-cam: Visual explanations from deep networks via gradient-based
  localization.
\newblock In \emph{{ICCV}}, pages 618--626. 2017.

\bibitem{VL-BERT}
Su, W., X.~Zhu, Y.~Cao, et~al.
\newblock Vl-bert: Pre-training of generic visual-linguistic representations.
\newblock In \emph{{ICLR}}. 2020.

\bibitem{ViLBERT}
Lu, J., D.~Batra, D.~Parikh, et~al.
\newblock Vilbert: Pretraining task-agnostic visiolinguistic representations
  for vision-and-language tasks.
\newblock In H.~M. Wallach, H.~Larochelle, A.~Beygelzimer,
  F.~d'Alch{\'{e}}{-}Buc, E.~B. Fox, R.~Garnett, eds., \emph{NeurIPS}, pages
  13--23. 2019.

\bibitem{12in1}
Lu, J., V.~Goswami, M.~Rohrbach, et~al.
\newblock 12-in-1: Multi-task vision and language representation learning.
\newblock In \emph{{CVPR}}, pages 10434--10443. 2020.

\bibitem{VisualBERT}
Li, L.~H., M.~Yatskar, D.~Yin, et~al.
\newblock Visualbert: {A} simple and performant baseline for vision and
  language.
\newblock \emph{arXiv preprint arXiv:1908.03557}, abs/1908.03557, 2019.

\bibitem{ImageBERT}
Qi, D., L.~Su, J.~Song, et~al.
\newblock Imagebert: Cross-modal pre-training with large-scale weak-supervised
  image-text data.
\newblock \emph{arXiv preprint arXiv:2001.07966}, 2020.

\bibitem{unicoder}
Li, G., N.~Duan, Y.~Fang, et~al.
\newblock Unicoder-vl: {A} universal encoder for vision and language by
  cross-modal pre-training.
\newblock In \emph{{AAAI}}, pages 11336--11344. 2020.

\bibitem{ernie}
Yu, F., J.~Tang, W.~Yin, et~al.
\newblock Ernie-vil: Knowledge enhanced vision-language representations through
  scene graph.
\newblock \emph{arXiv preprint arXiv:2006.16934}, 2020.

\bibitem{vinvl}
Zhang, P., X.~Li, X.~Hu, et~al.
\newblock Vinvl: Making visual representations matter in vision-language
  models.
\newblock \emph{arXiv preprint arXiv:2101.00529}, 2021.

\bibitem{soho}
Huang, Z., Z.~Zeng, Y.~Huang, et~al.
\newblock Seeing out of the box: End-to-end pre-training for vision-language
  representation learning.
\newblock \emph{arXiv preprint arXiv:2104.03135}, 2021.

\bibitem{NLVR}
Suhr, A., S.~Zhou, A.~Zhang, et~al.
\newblock A corpus for reasoning about natural language grounded in
  photographs.
\newblock In A.~Korhonen, D.~R. Traum, L.~M{\`{a}}rquez, eds., \emph{{ACL}},
  pages 6418--6428. 2019.

\bibitem{VQA}
Antol, S., A.~Agrawal, J.~Lu, et~al.
\newblock {VQA:} visual question answering.
\newblock In \emph{{ICCV}}, pages 2425--2433. 2015.

\bibitem{ViLT}
Kim, W., B.~Son, I.~Kim.
\newblock Vilt: Vision-and-language transformer without convolution or region
  supervision.
\newblock \emph{arXiv preprint arXiv:2102.03334}, 2021.

\bibitem{VSE}
Faghri, F., D.~J. Fleet, J.~R. Kiros, et~al.
\newblock {VSE++:} improving visual-semantic embeddings with hard negatives.
\newblock In \emph{{BMVC}}, page~12. 2018.

\bibitem{VSR}
Li, K., Y.~Zhang, K.~Li, et~al.
\newblock Visual semantic reasoning for image-text matching.
\newblock In \emph{{ICCV}}, pages 4653--4661. 2019.

\bibitem{moco}
He, K., H.~Fan, Y.~Wu, et~al.
\newblock Momentum contrast for unsupervised visual representation learning.
\newblock In \emph{CVPR}. 2020.

\bibitem{simclr}
Chen, T., S.~Kornblith, M.~Norouzi, et~al.
\newblock A simple framework for contrastive learning of visual
  representations.
\newblock In \emph{{ICML}}. 2020.

\bibitem{PCL}
Li, J., P.~Zhou, C.~Xiong, et~al.
\newblock Prototypical contrastive learning of unsupervised representations.
\newblock In \emph{ICLR}. 2021.

\bibitem{mopro}
Li, J., C.~Xiong, S.~C. Hoi.
\newblock Mopro: Webly supervised learning with momentum prototypes.
\newblock In \emph{ICLR}. 2021.

\bibitem{kd_hinton}
Hinton, G., O.~Vinyals, J.~Dean.
\newblock Distilling the knowledge in a neural network.
\newblock \emph{arXiv preprint arXiv:1503.02531}, 2015.

\bibitem{kd_att}
Zagoruyko, S., N.~Komodakis.
\newblock Paying more attention to attention: Improving the performance of
  convolutional neural networks via attention transfer.
\newblock In \emph{{ICLR}}. 2017.

\bibitem{born_again}
Furlanello, T., Z.~C. Lipton, M.~Tschannen, et~al.
\newblock Born-again neural networks.
\newblock In J.~G. Dy, A.~Krause, eds., \emph{{ICML}}, pages 1602--1611. 2018.

\bibitem{deit}
Touvron, H., M.~Cord, M.~Douze, et~al.
\newblock Training data-efficient image transformers \& distillation through
  attention.
\newblock \emph{arXiv preprint arXiv:2012.12877}, 2020.

\bibitem{distilbert}
Sanh, V., L.~Debut, J.~Chaumond, et~al.
\newblock Distilbert, a distilled version of bert: smaller, faster, cheaper and
  lighter.
\newblock \emph{arXiv preprint arXiv:1910.01108}, 2019.

\bibitem{DML}
Zhang, Y., T.~Xiang, T.~M. Hospedales, et~al.
\newblock Deep mutual learning.
\newblock In \emph{{CVPR}}, pages 4320--4328. 2018.

\bibitem{co_distill}
Anil, R., G.~Pereyra, A.~Passos, et~al.
\newblock Large scale distributed neural network training through online
  distillation.
\newblock In \emph{{ICLR}}. 2018.

\bibitem{mean_teacher}
Tarvainen, A., H.~Valpola.
\newblock Mean teachers are better role models: Weight-averaged consistency
  targets improve semi-supervised deep learning results.
\newblock In \emph{{NIPS}}, pages 1195--1204. 2017.

\bibitem{dividemix}
Li, J., R.~Socher, S.~C. Hoi.
\newblock Dividemix: Learning with noisy labels as semi-supervised learning.
\newblock In \emph{{ICLR}}. 2020.

\bibitem{self_distill}
Cheng, R., B.~Wu, P.~Zhang, et~al.
\newblock Data-efficient language-supervised zero-shot learning with
  self-distillation.
\newblock \emph{arXiv preprint arXiv:2104.08945}, 2021.

\bibitem{vit}
Dosovitskiy, A., L.~Beyer, A.~Kolesnikov, et~al.
\newblock An image is worth 16x16 words: Transformers for image recognition at
  scale.
\newblock In \emph{ICLR}. 2021.

\bibitem{transformer}
Vaswani, A., N.~Shazeer, N.~Parmar, et~al.
\newblock Attention is all you need.
\newblock In I.~Guyon, U.~von Luxburg, S.~Bengio, H.~M. Wallach, R.~Fergus,
  S.~V.~N. Vishwanathan, R.~Garnett, eds., \emph{{NIPS}}, pages 5998--6008.
  2017.

\bibitem{bert}
Devlin, J., M.~Chang, K.~Lee, et~al.
\newblock {BERT:} pre-training of deep bidirectional transformers for language
  understanding.
\newblock In J.~Burstein, C.~Doran, T.~Solorio, eds., \emph{{NAACL}}, pages
  4171--4186. 2019.

\bibitem{coco}
Lin, T., M.~Maire, S.~J. Belongie, et~al.
\newblock Microsoft {COCO:} common objects in context.
\newblock In D.~J. Fleet, T.~Pajdla, B.~Schiele, T.~Tuytelaars, eds.,
  \emph{{ECCV}}, vol. 8693, pages 740--755. 2014.

\bibitem{VG}
Krishna, R., Y.~Zhu, O.~Groth, et~al.
\newblock Visual genome: Connecting language and vision using crowdsourced
  dense image annotations.
\newblock \emph{{IJCV}}, 123(1):32--73, 2017.

\bibitem{cc12m}
Changpinyo, S., P.~Sharma, N.~Ding, et~al.
\newblock {Conceptual 12M}: Pushing web-scale image-text pre-training to
  recognize long-tail visual concepts.
\newblock In \emph{CVPR}. 2021.

\bibitem{adamw}
Loshchilov, I., F.~Hutter.
\newblock Decoupled weight decay regularization.
\newblock \emph{arXiv preprint arXiv:1711.05101}, 2017.

\bibitem{randaug}
Cubuk, E.~D., B.~Zoph, J.~Shlens, et~al.
\newblock Randaugment: Practical automated data augmentation with a reduced
  search space.
\newblock In \emph{{CVPR} Workshops}, pages 702--703. 2020.

\bibitem{goodview}
Tian, Y., C.~Sun, B.~Poole, et~al.
\newblock What makes for good views for contrastive learning?
\newblock In H.~Larochelle, M.~Ranzato, R.~Hadsell, M.~Balcan, H.~Lin, eds.,
  \emph{NeurIPS}. 2020.

\bibitem{CPC}
Oord, A. v.~d., Y.~Li, O.~Vinyals.
\newblock Representation learning with contrastive predictive coding.
\newblock \emph{arXiv preprint arXiv:1807.03748}, 2018.

\bibitem{MI}
Kong, L., C.~de~Masson~d'Autume, L.~Yu, et~al.
\newblock A mutual information maximization perspective of language
  representation learning.
\newblock In \emph{{ICLR}}. OpenReview.net, 2020.

\bibitem{flickr}
Plummer, B.~A., L.~Wang, C.~M. Cervantes, et~al.
\newblock Flickr30k entities: Collecting region-to-phrase correspondences for
  richer image-to-sentence models.
\newblock In \emph{{ICCV}}, pages 2641--2649. 2015.

\bibitem{ve-2.0}
Do, V., O.-M. Camburu, Z.~Akata, et~al.
\newblock e-snli-ve: Corrected visual-textual entailment with natural language
  explanations.
\newblock \emph{arXiv preprint arXiv:2004.03744}, 2020.

\bibitem{VE}
Xie, N., F.~Lai, D.~Doran, et~al.
\newblock Visual entailment: A novel task for fine-grained image understanding.
\newblock \emph{arXiv preprint arXiv:1901.06706}, 2019.

\bibitem{VQA2}
Goyal, Y., T.~Khot, D.~Summers{-}Stay, et~al.
\newblock Making the {V} in {VQA} matter: Elevating the role of image
  understanding in visual question answering.
\newblock In \emph{{CVPR}}, pages 6325--6334. 2017.

\bibitem{MAttNet}
Yu, L., Z.~Lin, X.~Shen, et~al.
\newblock Mattnet: Modular attention network for referring expression
  comprehension.
\newblock In \emph{{CVPR}}, pages 1307--1315. 2018.

\bibitem{VL_T5}
Cho, J., J.~Lei, H.~Tan, et~al.
\newblock Unifying vision-and-language tasks via text generation.
\newblock \emph{arXiv preprint arXiv:2102.02779}, 2021.

\bibitem{bilinear}
Kim, J., J.~Jun, B.~Zhang.
\newblock Bilinear attention networks.
\newblock In S.~Bengio, H.~M. Wallach, H.~Larochelle, K.~Grauman,
  N.~Cesa{-}Bianchi, R.~Garnett, eds., \emph{{NIPS}}, pages 1571--1581. 2018.

\bibitem{refer}
Yu, L., P.~Poirson, S.~Yang, et~al.
\newblock Modeling context in referring expressions.
\newblock In B.~Leibe, J.~Matas, N.~Sebe, M.~Welling, eds., \emph{{ECCV}},
  pages 69--85. 2016.

\bibitem{ARN}
Liu, X., L.~Li, S.~Wang, et~al.
\newblock Adaptive reconstruction network for weakly supervised referring
  expression grounding.
\newblock In \emph{{ICCV}}, pages 2611--2620. 2019.

\bibitem{CCL}
Zhang, Z., Z.~Zhao, Z.~Lin, et~al.
\newblock Counterfactual contrastive learning fo weakly-supervised
  vision-language grounding.
\newblock In H.~Larochelle, M.~Ranzato, R.~Hadsell, M.~Balcan, H.~Lin, eds.,
  \emph{NeurIPS}. 2020.

\bibitem{karpathy}
Karpathy, A., F.~Li.
\newblock Deep visual-semantic alignments for generating image descriptions.
\newblock In \emph{{CVPR}}, pages 3128--3137. 2015.

\bibitem{SNLI}
Bowman, S.~R., G.~Angeli, C.~Potts, et~al.
\newblock A large annotated corpus for learning natural language inference.
\newblock In L.~M{\`{a}}rquez, C.~Callison{-}Burch, J.~Su, D.~Pighin,
  Y.~Marton, eds., \emph{{EMNLP}}, pages 632--642. 2015.

\bibitem{co_attention}
Yu, Z., J.~Yu, Y.~Cui, et~al.
\newblock Deep modular co-attention networks for visual question answering.
\newblock In \emph{{CVPR}}, pages 6281--6290. 2019.

\bibitem{referit}
Kazemzadeh, S., V.~Ordonez, M.~Matten, et~al.
\newblock Referitgame: Referring to objects in photographs of natural scenes.
\newblock In A.~Moschitti, B.~Pang, W.~Daelemans, eds., \emph{{EMNLP}}. 2014.

\bibitem{vqahat}
Das, A., H.~Agrawal, C.~L. Zitnick, et~al.
\newblock {Human Attention in Visual Question Answering: Do Humans and Deep
  Networks Look at the Same Regions?}
\newblock 2016.

\end{thebibliography}

\newpage

\appendix

\vspace{-1.5ex}
\section{Downstream Task Details}
\vspace{-1.5ex}
Here we describe the implementation details for fine-tuning the pre-trained model.
For all downstream tasks,
we use the same RandAugment, AdamW optimizer, cosine learning rate decay, weight decay, and distillation weight as during pre-training.
All downstream tasks receive input images of resolution $384\times384$.
During inference,
we resize the images without any cropping.

\noindent\textbf{Image-Text Retrieval.}
We consider two datasets for this task: COCO and Flickr30K.
We adopt the widely used Karpathy split~\cite{karpathy} for both datasets.
COCO contains 113/5k/5k for train/validation/test.
Flickr30K contains 29k/1k/1k images for train/validation/test.
We fine-tune for 10 epochs.
The batch size is 256 and the initial learning rate is $1e^{-5}$.
%The test temperature $\tau_\mathrm{test}$ is set as 0.4 based on validation performance.

\noindent\textbf{Visual Entailment.}
We evaluate on the SNLI-VE dataset~\cite{VE},
which is constructed using the Stanford Natural Language Inference (SNLI)~\cite{SNLI} and Flickr30K datasets.
We follow the original dataset split with 29.8k images for training, 
1k for evaluation, and 1k for test.
We fine-tune the pre-trained model for 5 epochs with a batch size of 256 and an initial learning rate of $2e^{-5}$.

\noindent\textbf{VQA.}
We conduct experiment on the VQA2.0 dataset~\cite{VQA2}, which is constructed using images from COCO.
It contains 83k images for training, 41k for validation, and 81k for test.
We report performance on the test-dev and test-std splits.
Following most existing works~\cite{LXMERT, uniter,co_attention},
we use both training and validation sets for training,
and include additional question-answer pairs from Visual Genome.
Because many questions in the VQA dataset contains multiple answers,
we weight the loss for each answer by its percentage of occurrence among all answers.
We fine-tune the model for 8 epochs,
using a batch size of 256 and an initial learning rate of $2e^{-5}$.

\noindent\textbf{NLVR$^2$.}
We conduct experiments following the original train/val/test split in~\cite{NLVR}.
We fine-tune the model for 10 epochs,
using a batch size of 128 and an initial learning rate of $2e^{-5}$.
Because NLVR receives two input images,
we perform an additional step of pre-training with text-assignment (TA) to prepare the model for reasoning over two images.
The TA pre-training uses images of size $256\times256$.
We pre-train for 1 epoch on the 4M dataset, using a batch size of 256 and a learning rate of $2e^{-5}$.

\begin{figure*}[h]
	\vspace{-0.8ex}
	\centering
	\includegraphics[width=\textwidth]{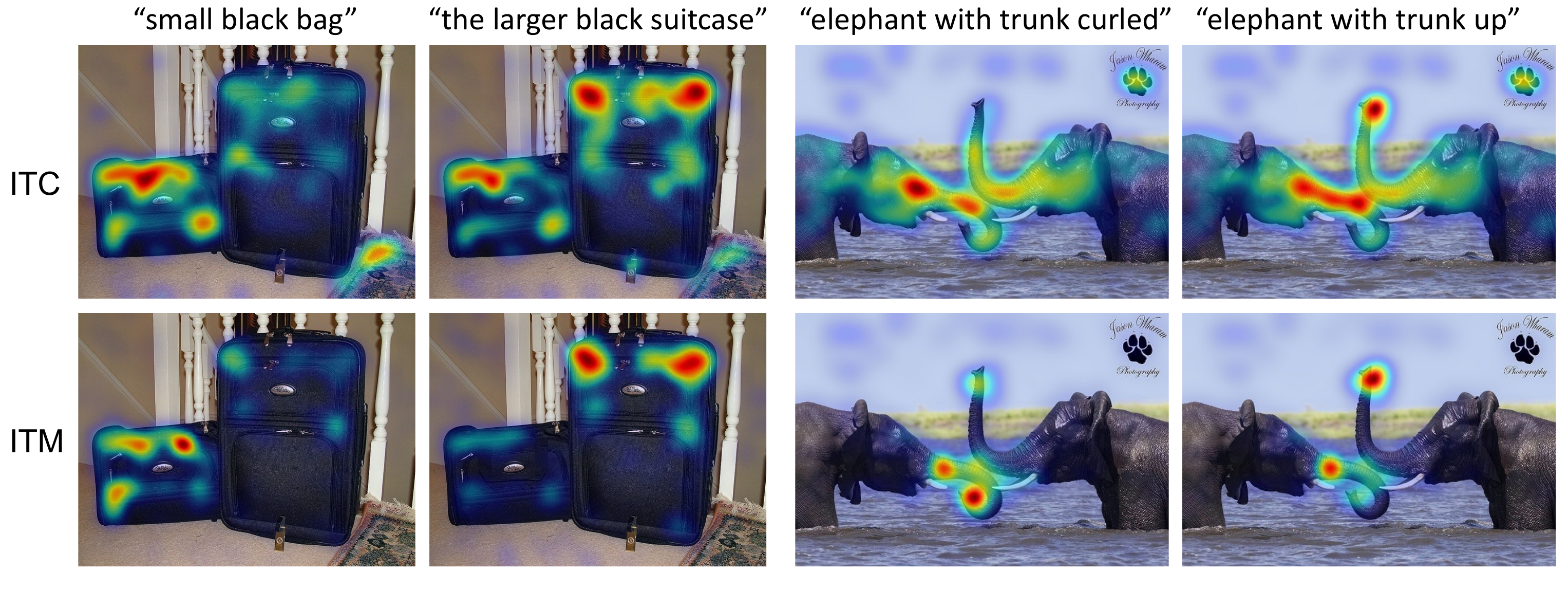}
	\vspace{-4.8ex}
	\caption
	{ \small
		Grad-CAMs from the multimodal encoder capture finer-grained details such as ``larger'' and ``curled''.
		} 
	\label{fig:gradcam_compare}
	\vspace{-1ex}
\end{figure*}

\begin{figure*}[!t]
	\vspace{-0.5ex}
	\centering
	\begin{minipage}{0.465\textwidth}
	\centering
	\includegraphics[width=\textwidth,trim={0 0.1cm 0 0},clip] {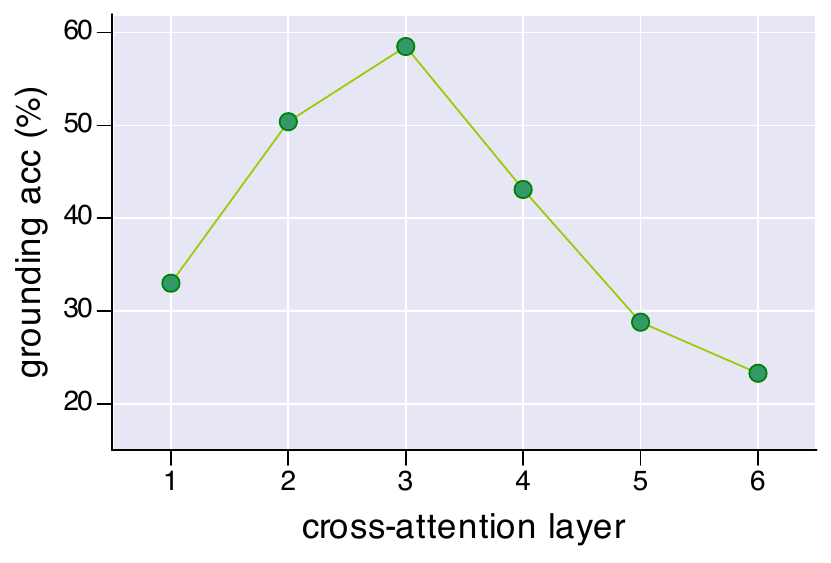}
	\vspace{-0.5ex}
	(a)
	\end{minipage}
	\hfill
	\begin{minipage}{0.50\textwidth}
	\centering
	\includegraphics[width=\textwidth,trim={0 0.1cm 0 0},clip]{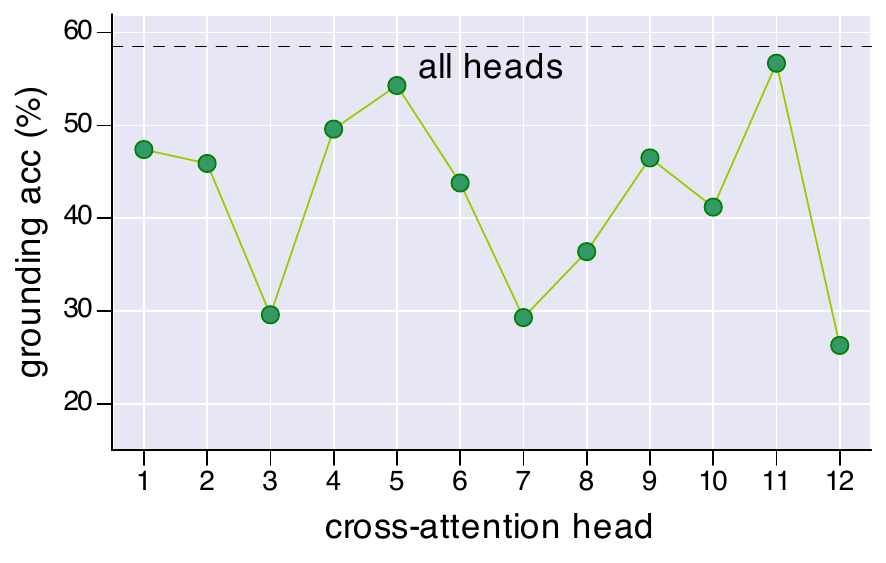}
	\vspace{-0.5ex}
	(b)
	\end{minipage}	
	\caption
	{ \small
		Grounding accuracy on the validation set of RefCOCO+.
		(a) varying cross-attention layers where each layer uses all heads.
		(b) varying cross-attention heads in the best-performing (3rd) layer.
		} 
	\label{fig:layer_head}
\end{figure*}

\noindent\textbf{Visual Grounding.}
We conduct experiments on the RefCOCO+ dataset~\cite{refer},
which is collected using a two-player ReferitGame~\cite{referit}.
It contains 141,564 expressions for 19,992 images from COCO training set.
Strictly speaking,
our model is not allowed to see the val/test images of RefCOCO+,
but it has been exposed to those images during pre-training.
We hypothesize that this has little effect because these images only occupy a very small portion of the entire 14M pre-training images,
and leave it as future work to decontaminate the data.
During weakly-supervised fine-tuning, we follow the same strategy as image-text retrieval except that we do not perform random cropping, and train the model for 5 epochs.
During inference, we use either $s_\mathrm{itc}$ or $s_\mathrm{itm}$ to compute the importance score for each $16\times16$ image patch.
For ITC,
we compute Grad-CAM visualizations on the self-attention maps \textit{w.r.t} the \texttt{[CLS]} token in the last layer of the visual encoder,
and average the heatmaps across all attention heads.
For ITM, we compute Grad-CAM on the cross-attention maps in the 3rd layer of the multimodal encoder, and average them scores across all attention heads and all input text tokens.
Quantitative comparison between ITC and ITM is shown in Table~\ref{tbl:grounding}.
Figure~\ref{fig:gradcam_compare} shows the qualitative comparison.
Since the multimodal encoder can better model image-text interactions, 
it produces better heatmaps that capture finer-grained details.
In Figure~\ref{fig:layer_head},
we report the grounding accuracy for each cross-attention layer
and each individual attention head within the best-performing layer.
% For future work,
% we intend to explore better methods for model interpretation beyond the simple averaging of GradCAMs.

\begin{figure*}[!t]
	\centering
	\includegraphics[width=\textwidth]{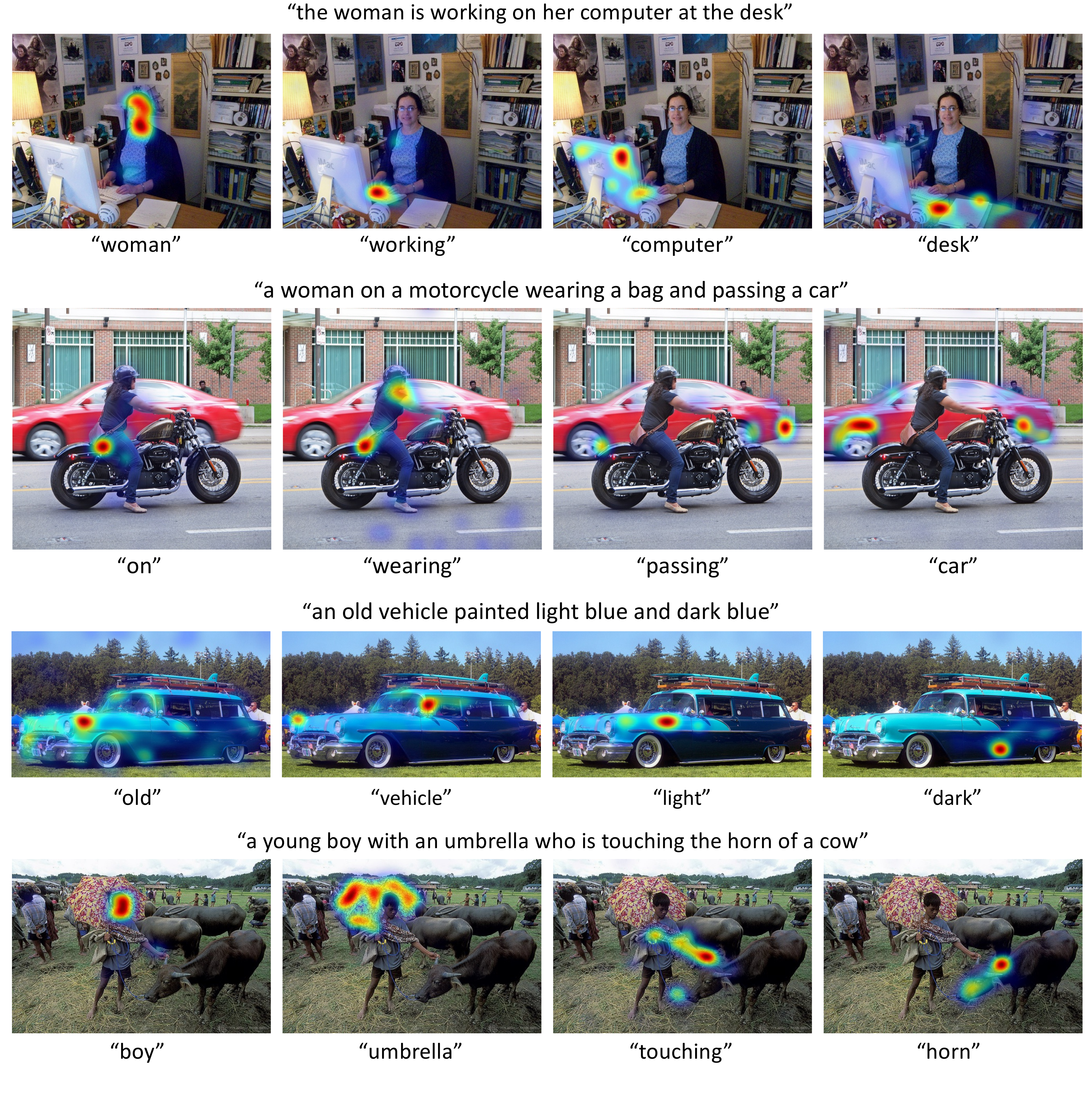}
 \vspace{-5ex}
  \caption
  	{ \small
		Grad-CAM visualization on the cross-attention maps corresponding to individual words.
	  } 
  \label{fig:coco_gradcam}
   \vspace{-4ex}
 \end{figure*} 
 
\vspace{-1ex}
\section{Additional Per-word Visualizations}
%\vspace{-1ex}
In Figure~\ref{fig:coco_gradcam},
we show more visualizations of per-word Grad-CAM to demonstrate the ability of our model to perform visual grounding of objects, actions, attributes, and relationships.

\vspace{-1.5ex}
\section{Comparison with Human Attention}
\begin{figure*}[!t]
	\centering
	\includegraphics[width=0.82\textwidth]{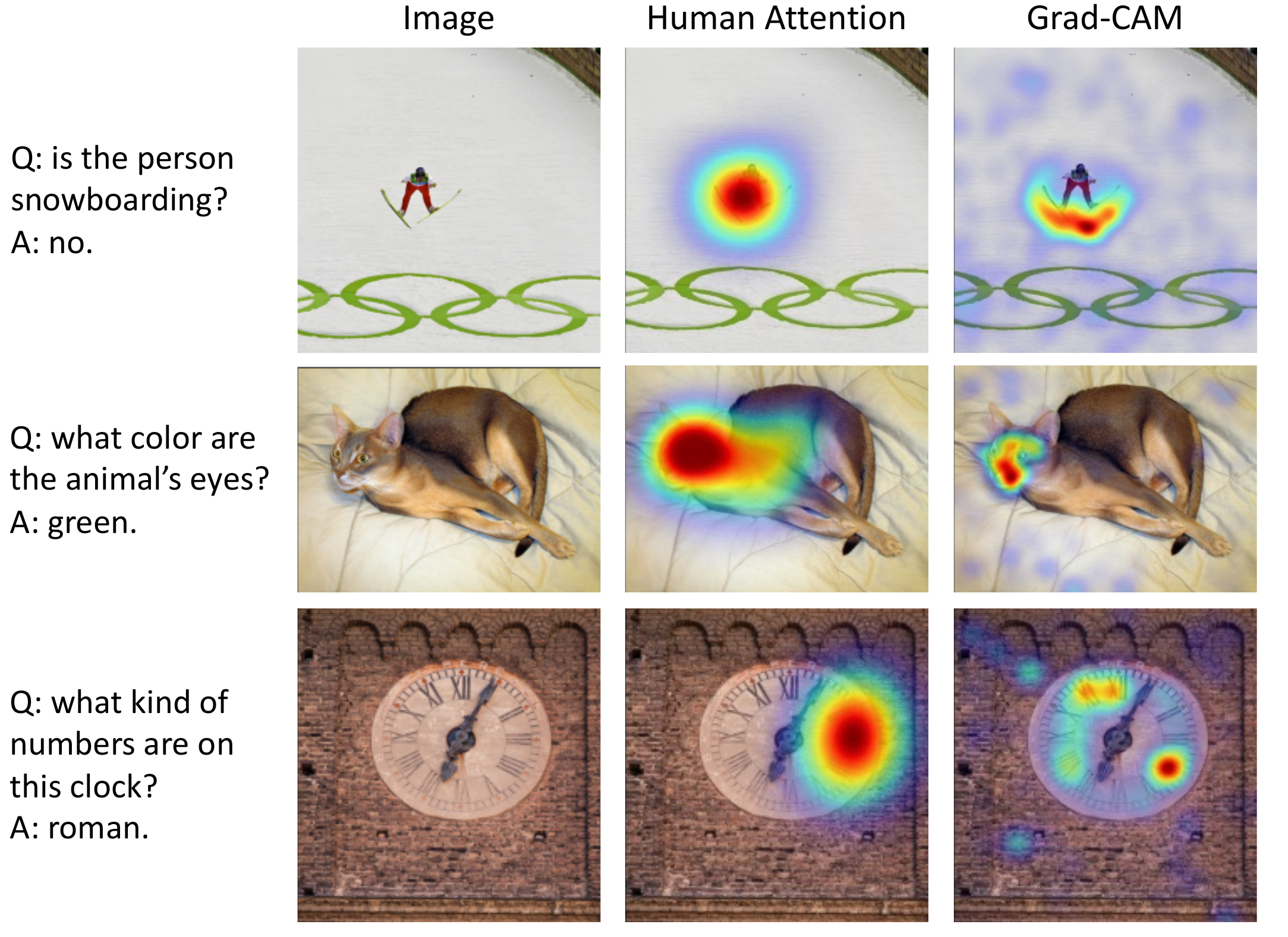}
 %\vspace{-2ex}
  \caption
  	{ \small
		Qualitative comparison between human attention and ALBEF's Grad-CAM for VQA.
	  } 
  \label{fig:vqa_gradcam}
   \vspace{-1ex}
 \end{figure*}

\vspace{-1ex}
Das \etal~\cite{vqahat} collected human attention maps for a subset of the VQA  dataset~\cite{VQA}. 
Given a question and a blurred version of the image, humans on Amazon Mechanical Turk were asked to interactively deblur image regions until they could confidently answer the question. 
In this work we compare human attention maps to Grad-CAM visualizations for the ALBEF VQA model computed at the 3rd multi-modal cross-attention layer on 1374 validation question-image pairs using the rank correlation evaluation protocol as in \cite{vqahat}. 
We find Grad-CAM and human attention maps computed for the ground-truth answer to have a high correlation of 0.205.
This shows that despite not being trained on grounded image-text pairs, ALBEF looks at appropriate regions when making decisions. 
Qualitative examples showing the comparison with human attention maps can be found in Figure~\ref{fig:vqa_gradcam}. 
 
\section{Additional Examples of Pseudo-targets}
\vspace{-1ex}
\begin{figure*}[!h]
	
	\centering
	\includegraphics[width=\textwidth]{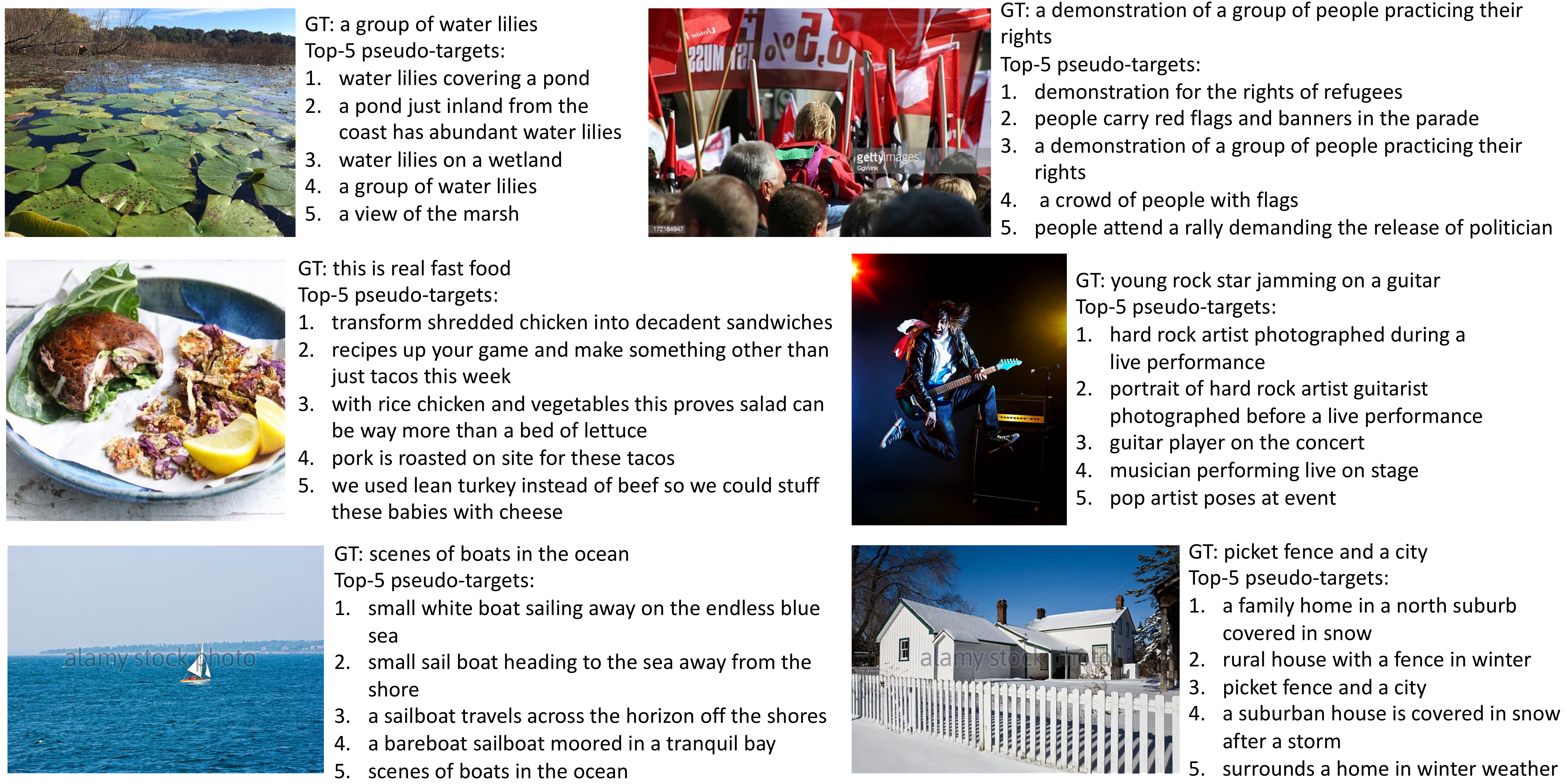}
	\caption
	{ \small
		Examples of the top-5 most similar texts selected by the momentum model for ITC.
	} 
	\label{fig:example_itc}
\end{figure*} 

\begin{figure*}[!h]
	
	\centering
	\includegraphics[width=\textwidth]{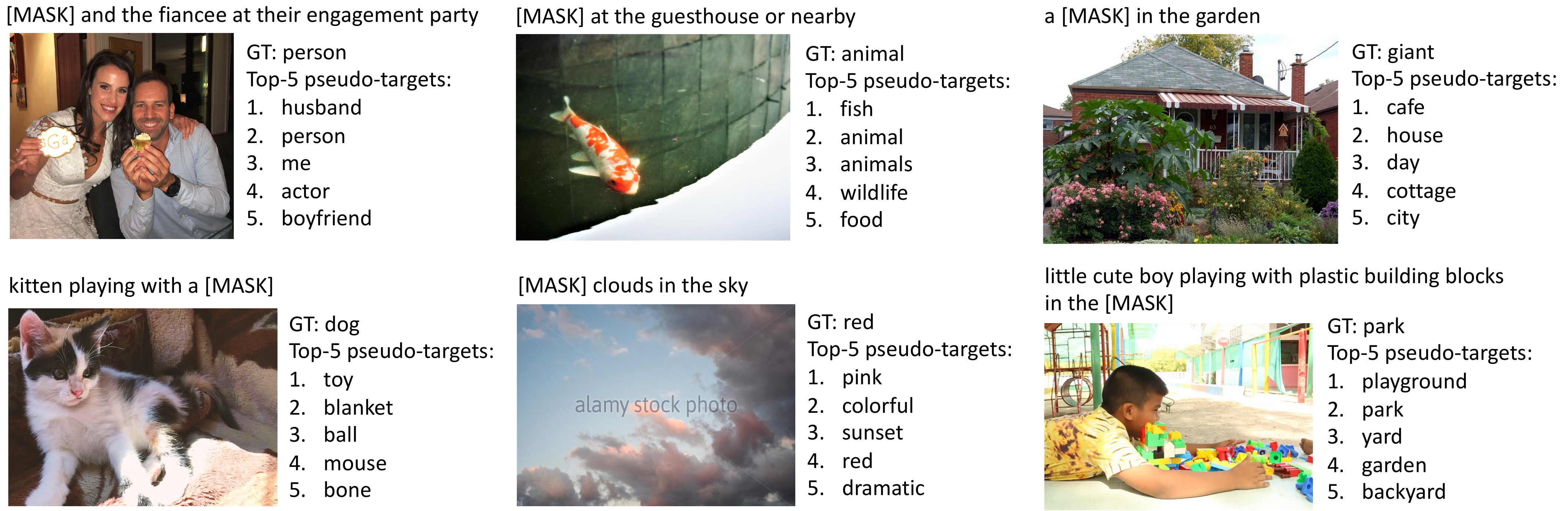}
	\caption
	{ \small
		Examples of the top-5 words generated by the momentum model for MLM.
	} 
	\label{fig:example_mlm}
\end{figure*} 

\section{Pre-training Dataset Details}

Table~\ref{tbl:pre-train_data} shows the statistics of the image and text of the pre-training datasets.
\begin{table*}[h]
	\small
	\centering	
	%\setlength\tabcolsep{3pt}
	%\resizebox{\textwidth}{!}{%
		\begin{tabular}	{l   |  c c c c c }
			\toprule
			  & COCO (Karpathy-train) & VG & CC & SBU& CC12M \\
			\midrule
			\# image  & 113K & 100K & 2.95M & 860K & 10.06M\\
			\# text & 567K & 769K & 2.95M & 860K & 10.06M\\
			\bottomrule
		\end{tabular}
	%}
	\caption
	{
		\small	
		Statistics of the pre-training datasets.
	}
	\label{tbl:pre-train_data}
\end{table*}

\end{document}